\documentclass[conference]{IEEEtran}
\IEEEoverridecommandlockouts

\usepackage{cite}
\usepackage{amsmath,amssymb,amsfonts}
\usepackage{algorithmic}
\usepackage{graphicx}
\usepackage{textcomp}
\usepackage{xcolor}
\usepackage{subcaption}

\usepackage{url}
\usepackage{hyperref}

\usepackage{amsmath}
\usepackage{amssymb}
\usepackage{booktabs}

\def\BibTeX{{\rm B\kern-.05em{\sc i\kern-.025em b}\kern-.08em
    T\kern-.1667em\lower.7ex\hbox{E}\kern-.125emX}}
\begin{document}

\title{CUEING: a lightweight model to Capture hUman attEntion In driviNG
}

\author{\IEEEauthorblockN{Linfeng Liang}
\IEEEauthorblockA{\textit{School of Computing} \\
\textit{Macquarie University}\\
Sydney, Australia \\
linfeng.liang@hdr.mq.edu.au}
\and
\IEEEauthorblockN{Yao Deng}
\IEEEauthorblockA{\textit{School of Computing} \\
\textit{Macquarie University}\\
Sydney, Australia \\
Yao.Deng@hdr.mq.edu.au}
\and
\IEEEauthorblockN{Yang Zhang}
\IEEEauthorblockA{\textit{School of Computing} \\
\textit{Macquarie University}\\
Sydney, Australia \\
yang.zhang@mq.edu.au}
\and
\IEEEauthorblockN{Jianchao Lu}
\IEEEauthorblockA{\textit{School of Computing} \\
\textit{Macquarie University}\\
Sydney, Australia \\
jianchao.lu@mq.edu.au}
\and
\IEEEauthorblockN{Chen Wang}
\IEEEauthorblockA{\textit{DATA61} \\
Sydney, Australia \\
chen.wang@data61.csiro.au}
\and
\IEEEauthorblockN{Quanzheng Sheng}
\IEEEauthorblockA{\textit{School of Computing} \\
\textit{Macquarie University}\\
Sydney, Australia \\
michael.sheng@mq.edu.au}
\and
\IEEEauthorblockN{Xi Zheng\IEEEauthorrefmark{1}\thanks{$^*$Corresponding author: james.zheng@mq.edu.au}}
\IEEEauthorblockA{\textit{School of Computing} \\
\textit{Macquarie University}\\
Sydney, Australia \\
james.zheng@mq.edu.au}
}





\maketitle

\begin{abstract}

Discrepancies in decision-making between Autonomous Driving Systems (ADS) and human drivers underscore the need for intuitive human gaze predictors to bridge this gap, 
thereby 
improving user trust and experience. 
Existing gaze datasets, despite their value, suffer from noise that hampers effective training. Furthermore, current gaze prediction models exhibit inconsistency across diverse scenarios and demand substantial computational resources, restricting their on-board deployment in autonomous vehicles. We propose a novel adaptive cleansing technique for purging noise from existing gaze datasets, coupled with a robust, lightweight convolutional self-attention gaze prediction model. Our approach not only significantly enhances model generalizability and performance by up to 12.13\%\ but also ensures a remarkable reduction in model complexity by up to 98.2\% compared to the state-of-the-art, making in-vehicle deployment feasible to augment ADS decision visualization and performance. 

 
\end{abstract}


\section{Introduction}
\label{sec:intro}

Recent studies show notable differences between autonomous driving systems and human drivers in decision-making, raising concerns about the reliability of such systems~\cite{accidentnews,news2}. Human drivers largely depend on visual assessments to react to road situations, particularly in hazardous scenarios. Recent research suggests that eye-tracking data from human drivers can accurately predict attention, hinting at the potential to enhance autonomous driving decision-making~\cite{codevilla2018end,xia2020periphery, zhang2021recent, bao2021drive, wang2023decision}. A lightweight human attention model emerges as a promising tool to bridge this gap. Deployed in self-driving vehicles, this model could provide real-time visual cues of autonomous decisions, enhancing user trust~\cite{bao2021drive, news3,news4,news5}. Additionally, coupled with eye-tracking devices, it could help in online training setups, making self-driving models more robust and human-centric \cite{news2}. Hence, developing such a model is crucial for advancing trustworthy autonomous driving systems.

\begin{figure}[!ht]
\centering
{\includegraphics[width=0.3\linewidth]{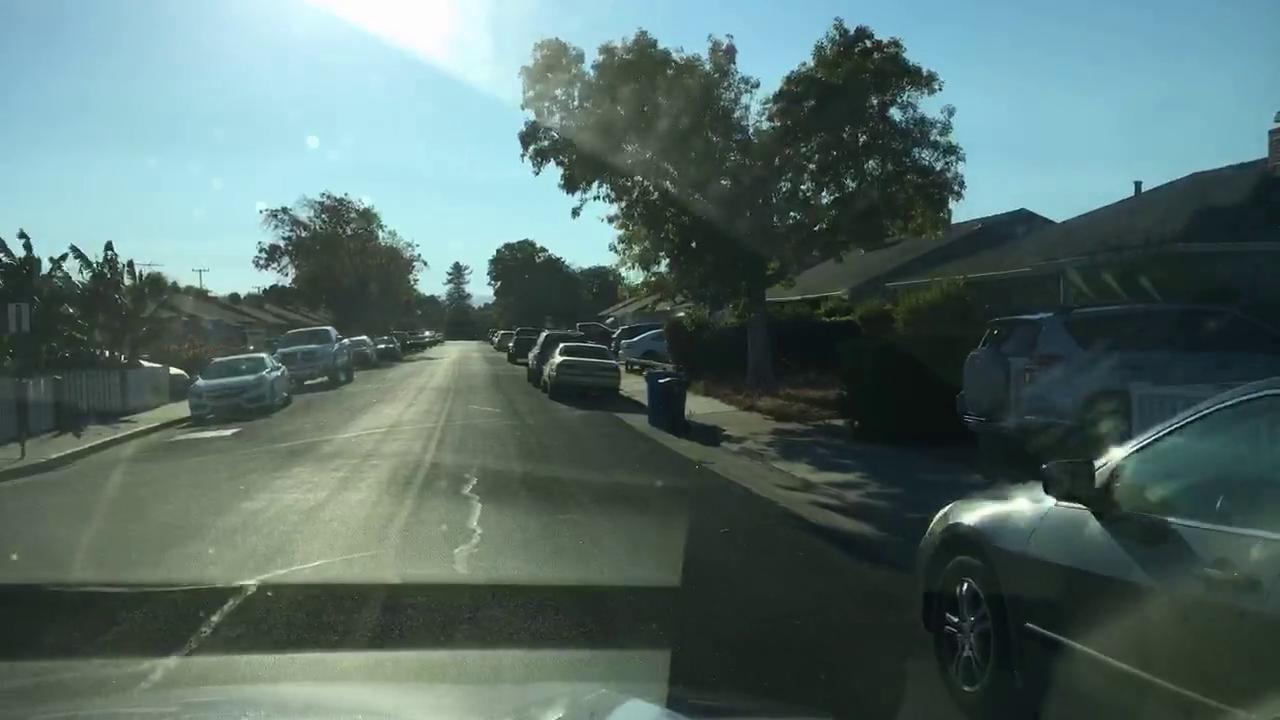}}
{\includegraphics[width=0.3\linewidth]{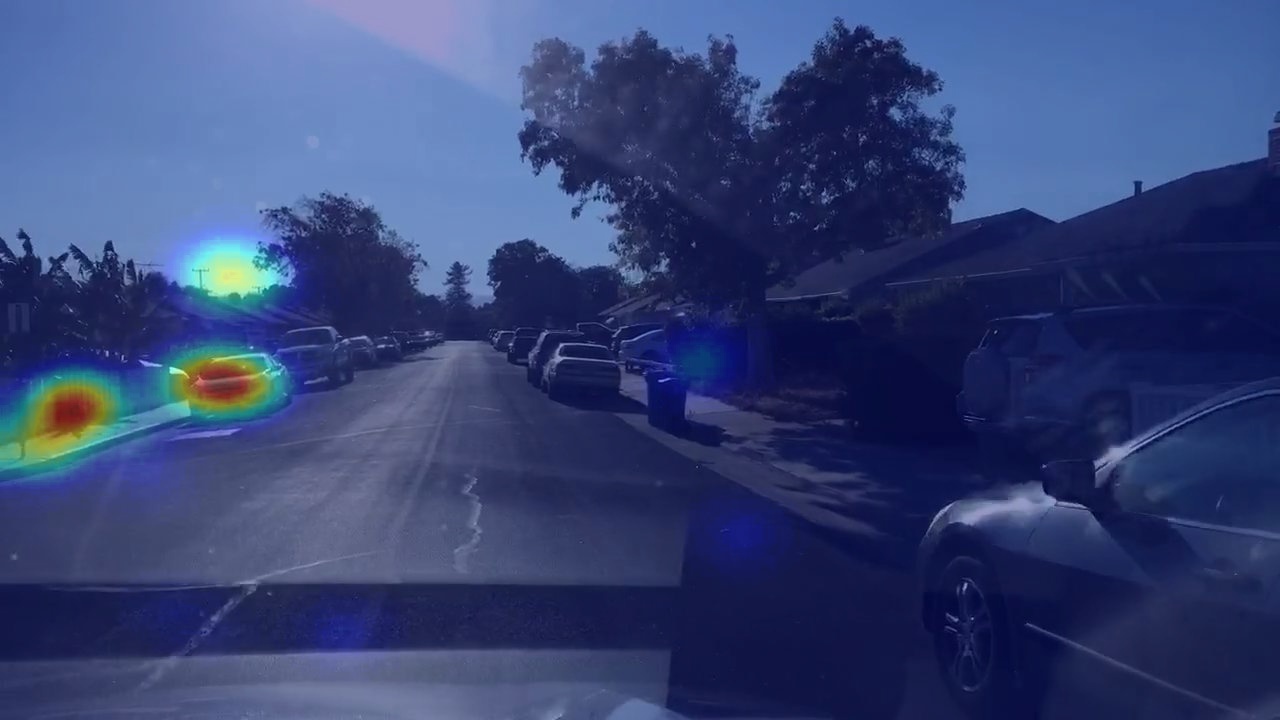}}
{\includegraphics[width=0.3\linewidth]{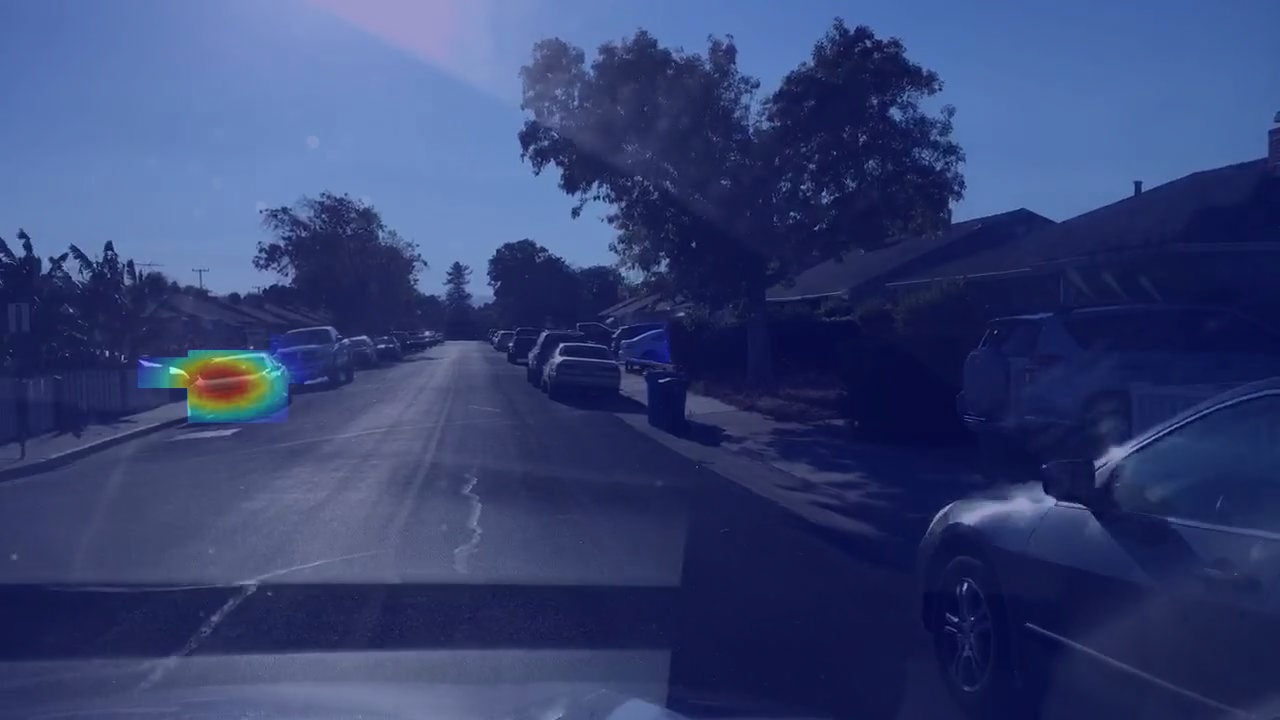}}

{\includegraphics[width=0.3\linewidth]{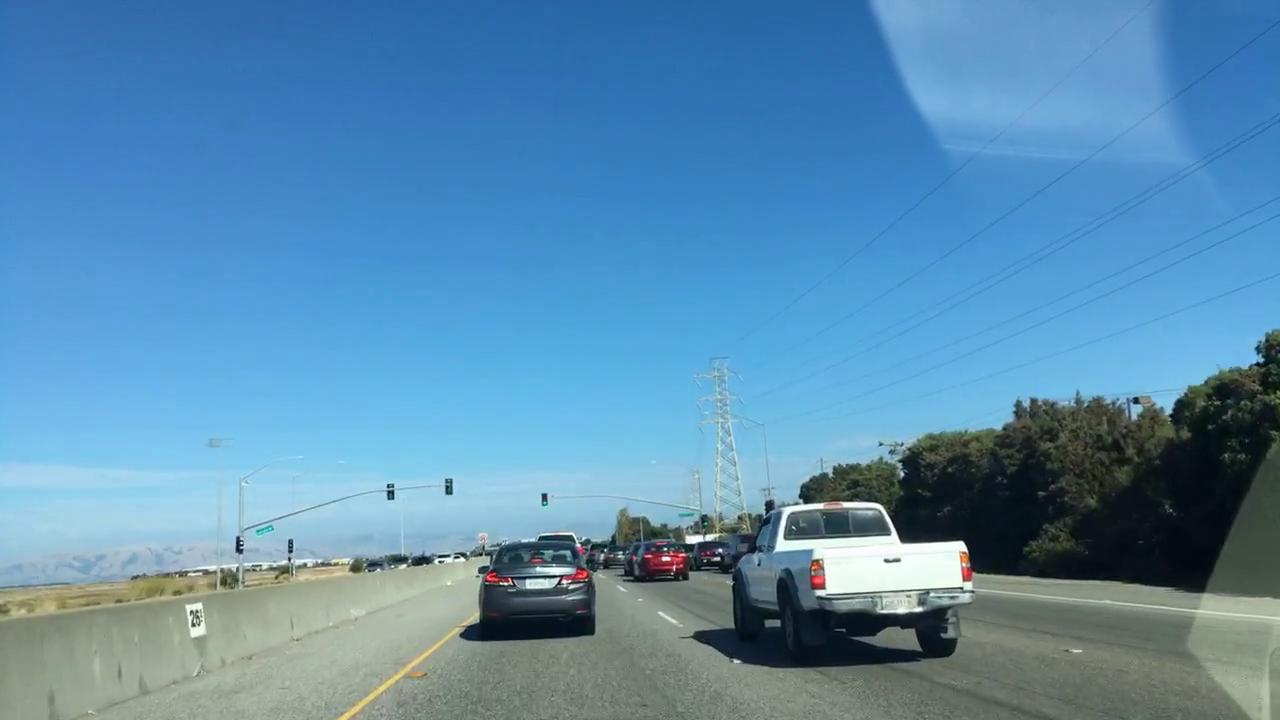}}
{\includegraphics[width=0.3\linewidth]{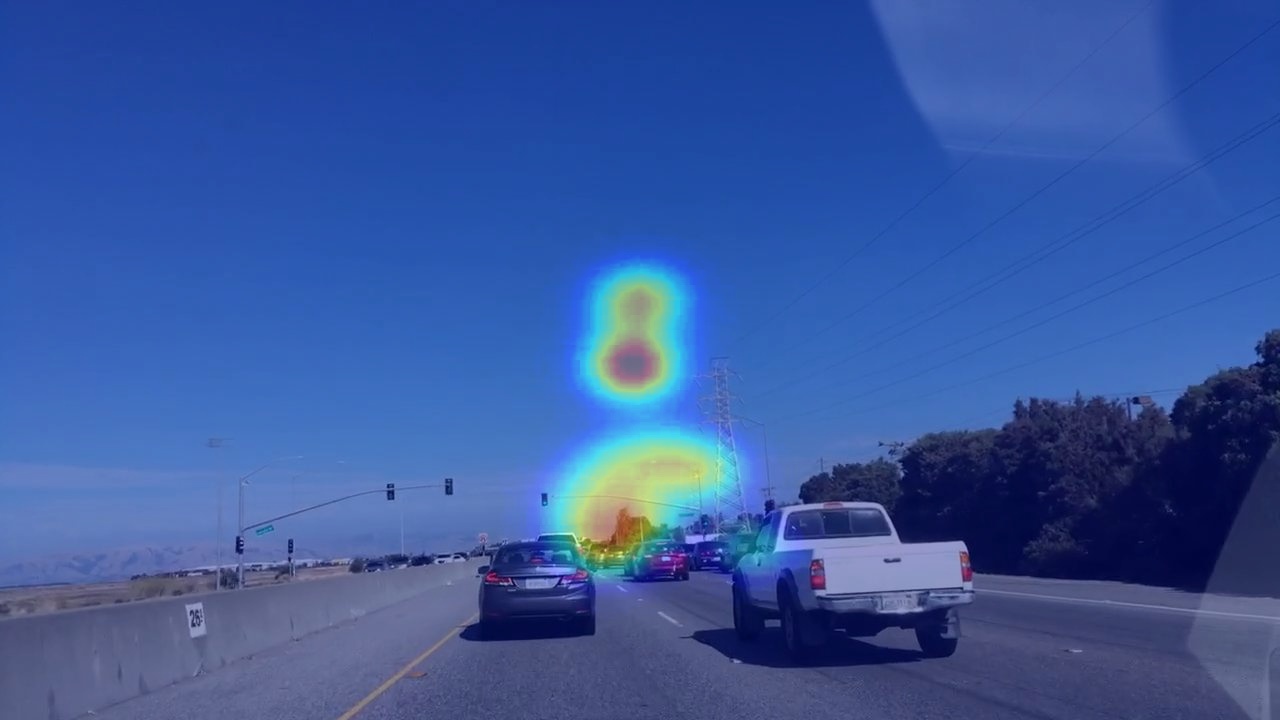}}
{\includegraphics[width=0.3\linewidth]{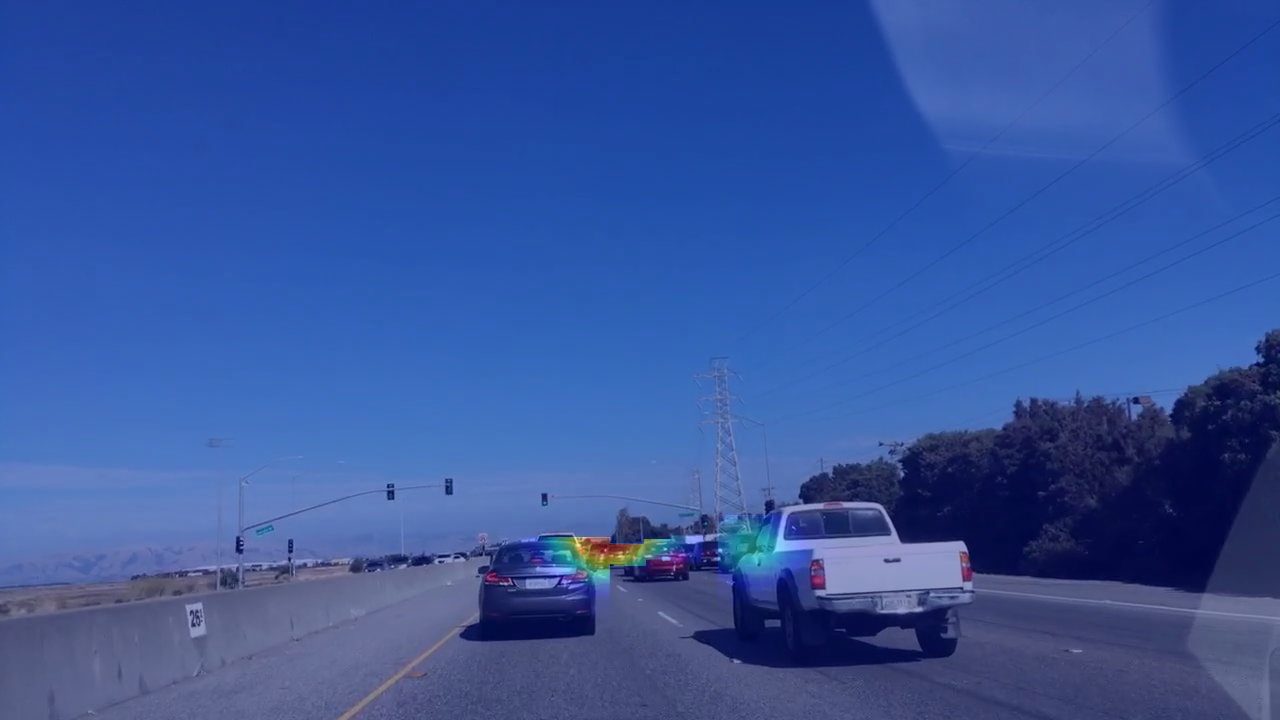}}

{\includegraphics[width=0.3\linewidth]{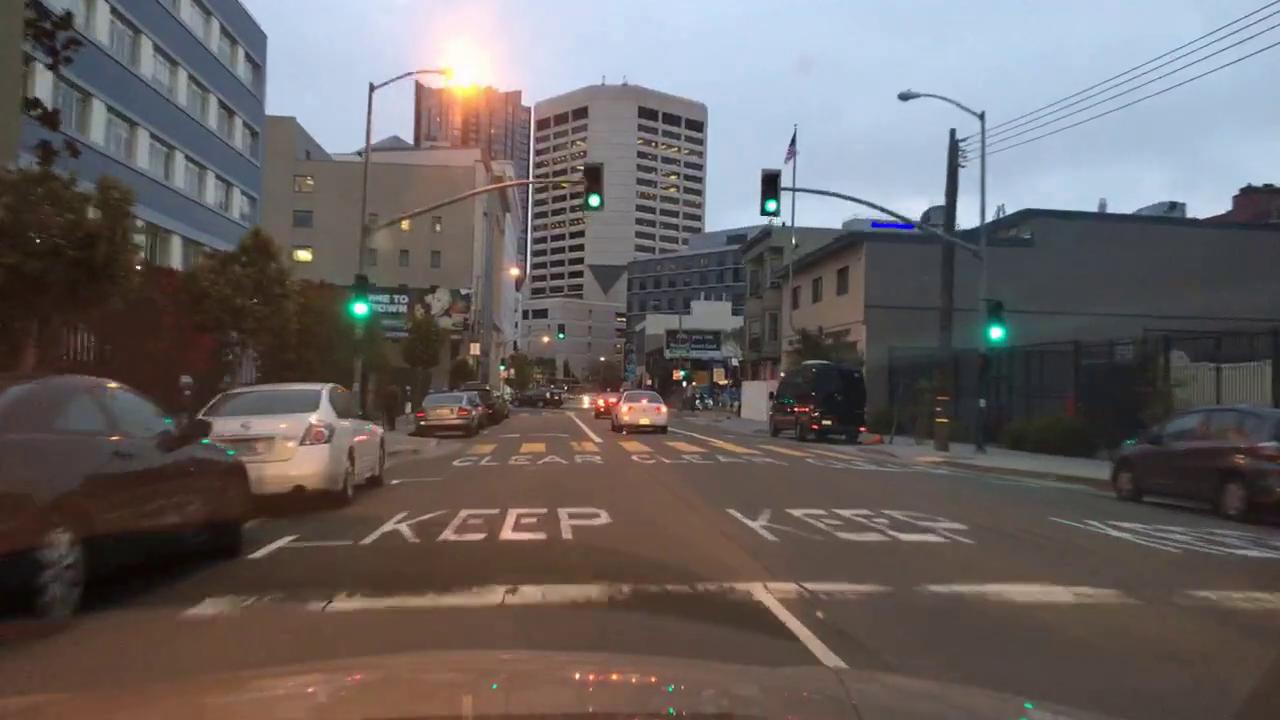}}
{\includegraphics[width=0.3\linewidth]{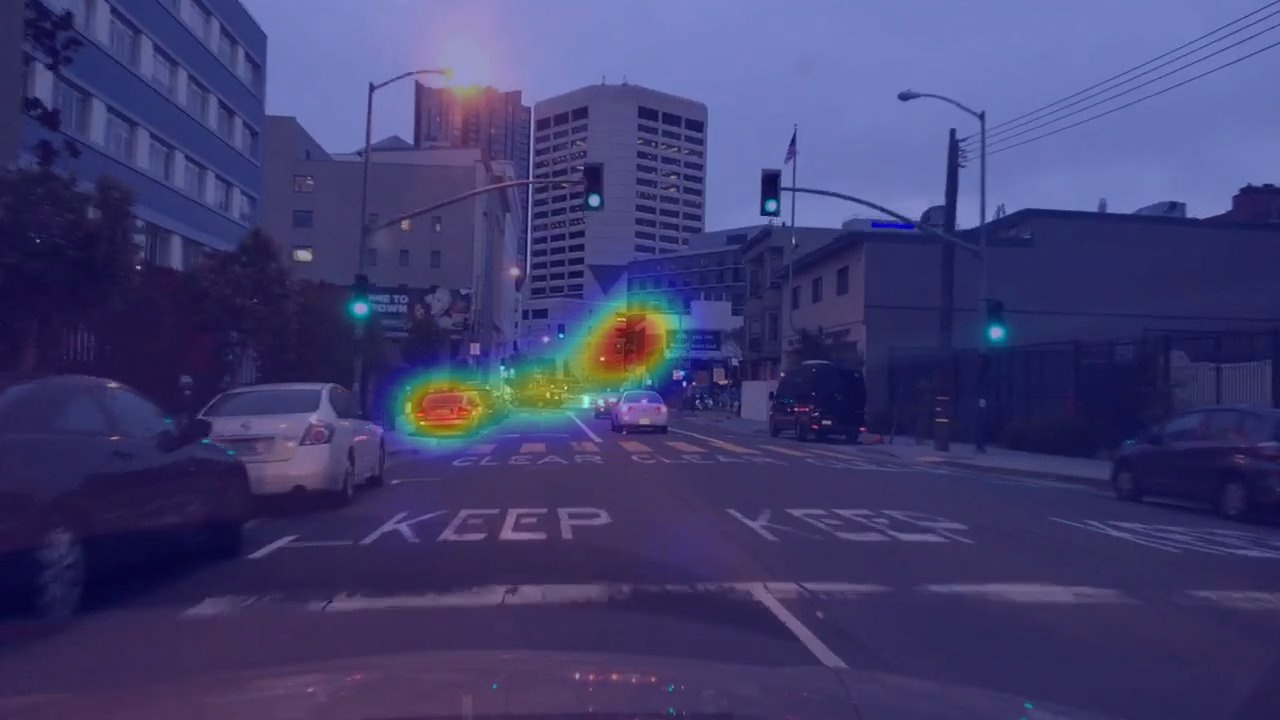}}
{\includegraphics[width=0.3\linewidth]{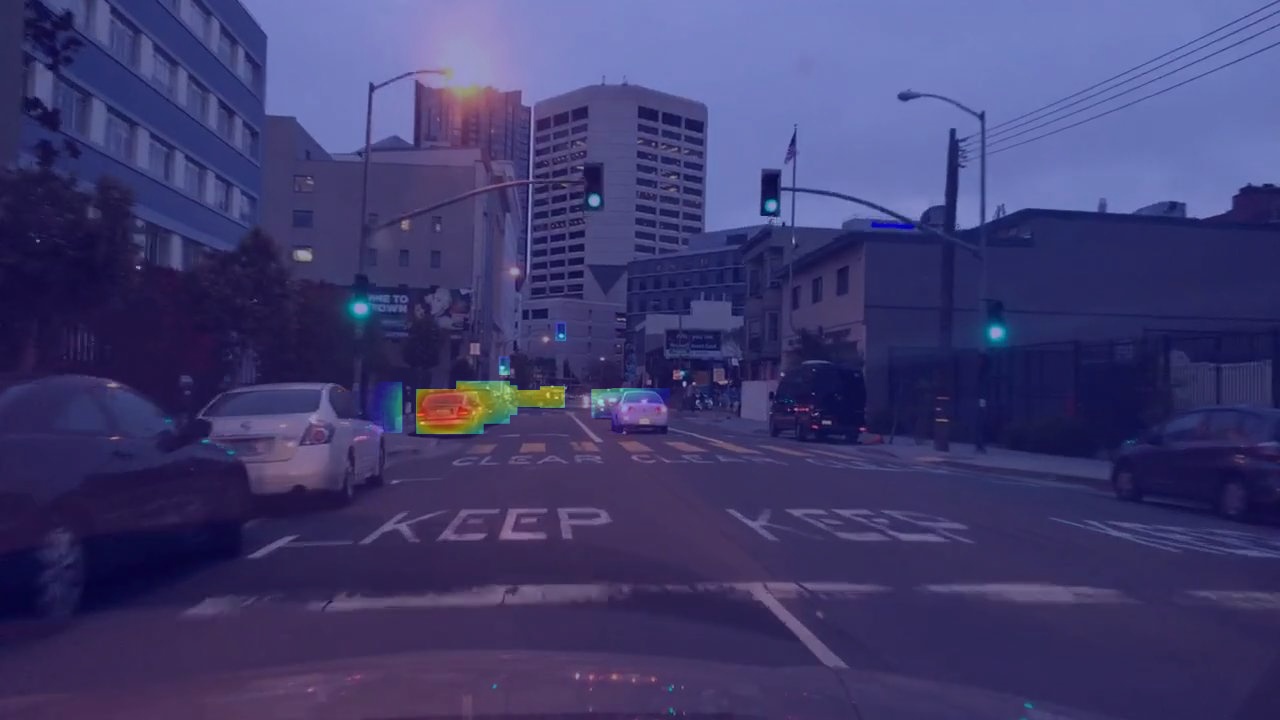}}
        
\caption{Camera images from BDD-A (first column), Gaze map from BDD-A (second column), Cleansed gaze map from BDD-A (third column).}

\label{TEST}
\end{figure}

However, despite the availability of extensive gaze data and the development of gaze prediction models \cite{xia2018predicting, fang2019dada, palazzi2018predicting, fang2021dada, rong2022and}, recent research findings suggest that the approach of using eye-tracking to capture drivers' attention poses certain challenges \cite{kotseruba2022attention, pal2020looking}. Specifically, it has been observed that human attention can be easily diverted towards irrelevant items during the driving process \cite{ahlstrom2021eye}, and such unrelated gaze data (\emph{Challenge 1}) can have a detrimental effect on the training of gaze prediction models. Figure \ref{TEST} indicates a comparison between the original gaze map in BDD-A \cite{xia2018predicting} (Column 2) and its cleansed version (Column 3). We can see that, although attempts have been made to collect the average gaze from multiple drivers \cite{xia2018predicting}, unrelated gaze (e.g., building, tree and poles) still widely exists in the BDD-A dataset compared to our cleansed version \cite{xia2018predicting}.

Furthermore, the gaze prediction models shall have good generalizability when applied to diverse driving scenarios. However, developing an effective gaze prediction model for a new environment often mandates the collection of extensive driving gaze data, which can be costly and time-consuming. Thus, another major challenge lies in the development of a more generalizable gaze prediction model that can perform well across diverse scenarios (\emph{Challenge 2}). Moreover, 
for
effective in-vehicle ADS decision-making assistance and to provide passengers with visual cues through gaze prediction, 
it is 
essential that the model is lightweight to ensure efficient computation on the resource-constrained devices deployed inside the vehicles
(\emph{Challenge 3}).

To address the aforementioned challenges, we propose a filtering process aimed at removing unrelated gaze data 
presented 
in current datasets, thus allowing the model to focus specifically on driving-related objects during the training process. Additionally, we design a convolutional self-attention model which boosts generalizability by evaluating the interrelations among tokens (regions) in the input images, and trims down model complexity through an efficient token convolution with a smaller kernel size and fewer channels.
%
In a nutshell, our major contributions are threefold:
\begin{itemize}
     \item We present an \textit{adaptive cleansing method} that employs the object's label and embedding information to generate masked inputs, facilitating the refinement of driving gaze datasets. Our cleansing method, when applied to  datasets, allows driving gaze prediction models to attain outstanding results.
    \item We propose a novel and \textit{light-weight convolutional self-attention gaze prediction model} that exhibits enhanced generalizability compared to existing gaze prediction models by employing tokenization and self-attention mechanisms.
      
    \item Our \textit{extensive experiments} reveal that our cleansing method improves model performance by as much as 8.75\%, while our attention model increases generalizability by up to 12.13\% over existing state-of-the-art models. Moreover, our model's parameters make up a mere 1.8\% of those in state-of-the-art models. 



\end{itemize}

The structure of this paper is outlined as follows. Section \ref{RW} delves into a comprehensive review of the pertinent literature related to our investigation. The creation of our dataset is elaborated in 
Section 
\ref{DC}, while 
Section 
\ref{CM} describes our modelling approach. The experimental design is highlighted in 
Section 
\ref{EX}. We 
report 
the outcomes of our rigorous experiments, rooted in the techniques detailed in 
Section 
\ref{DC},
\ref{CM}, 
and 
\ref{Result}. Lastly, potential vulnerabilities to our suggested approaches are 
discussed 
in 
Section 
 \ref{TV}.

\section{Related Work}
\label{RW}
\subsection{Driving Gaze Datasets}
Prior research has presented multiple driving gaze datasets. Dr(eye)VE \cite{alletto2016dr} consists of videos collected from real-world driving scenarios though with a smaller number of road users. BDD-A \cite{xia2018predicting} primarily focused on critical driving situations and employed in-lab methods to collect averaged gaze data from 20 drivers, which was gathered in San Francisco. Similarly, DADA-2000 \cite{fang2019dada} 
was 
generated using in-lab methods and 
comprised 
video clips depicting 54 different types of accidents across various environments. The dataset was collected in China, where cultural and
road conditions differ from previous datasets. Additionally, CDNN \cite{deng2019drivers} proposed a dataset that is collected in a similar way to DADA-2000 \cite{fang2019dada}. To counter the noise collected from eye-tracking equipment \cite{ahlstrom2021eye,kotseruba2022attention}, BDD-A \cite{xia2018predicting} used multiple drivers' averaged attention data as the ground truth in their work. 
However, as shown in Figure \ref{TEST}, some ground truth gaze still focuse on unrelated objects such as the sky or trees.
SAGE~\cite{pal2020looking} in another study improved label accuracy by adding objects' semantic segmentation to ground truth. It used Mask-RCNN to obtain objects' semantic segmentation such as vehicles and pedestrians.  Nonetheless, the resulting gaze map differs considerably from human gaze maps. 
In comparison, our cleansing method 
leverages 
the existing objects' labels and embedding information in the given dataset to extract the bounding boxes of those semantically relevant objects.
By masking all the other pixels, we 
are
 able to create a more accurate and focused gaze map. The experiments demonstrate that our gaze map successfully focuses on specific objects that are highly relevant to driving scenarios.

\subsection{Gaze Prediction Models}
Previous research primarily focuses on designing gaze prediction models using pre-trained CNN backbones, such as AlexNet \cite{xia2018predicting}, YoloV5 \cite{rong2022and}, or VGG \cite{simonyan2015deep}. This approach can be seen as mapping the extracted features to ground-truth gaze. Additionally, a grid-based gaze prediction method has been proposed, which divides the input into grids and predicts gaze for each grid \cite{rong2022and}. These methods have achieved decent quantitative results. Another approach, CDNN \cite{deng2019drivers}, provides a gaze predictor without a backbone but requires downsizing the original inputs. 
A major issue in these prior works is their heavy reliance on CNNs, which have limitations in capturing the relationships among different subfields in the input images. The gaze prediction task actually requires capturing these relationships effectively. Furthermore, previous models often require image pre-processing, either through feature extraction or downsizing.  Previous research \cite{9710860} also explored inverse reinforcement learning in gaze prediction. However, their method requires a range of prior labels involving relative distance, driving tasks, vehicle state, and road information, which can be challenging to obtain, especially across diverse datasets. 
Unlike the previous approach, ours provides a resource-efficient model. It avoids heavy reliance on convolution filters and pre-trained backbones. Instead, it uses a tokenization method to process the original input, utilizing self-attention to capture 
region relationships. Our model exhibits strong generalizability across various datasets.


\subsection{Model Compression}
High-performance deep learning models, characterized by their extensive parameters, consume considerable computational resources. To fit them for mobile deployment, downsizing is essential. While recent studies  \cite{celebi2015linear,han2015deep} suggest that model compression can reduce computational demands, it often results in diminished model performance. In our approach, we reduce the model's complexity and inference speed on resource-constrained devices with no compromise in accuracy.

\section{Dataset Cleansing}
\label{DC}

\begin{figure}[t]
    \centering
    \includegraphics[width=1\linewidth]{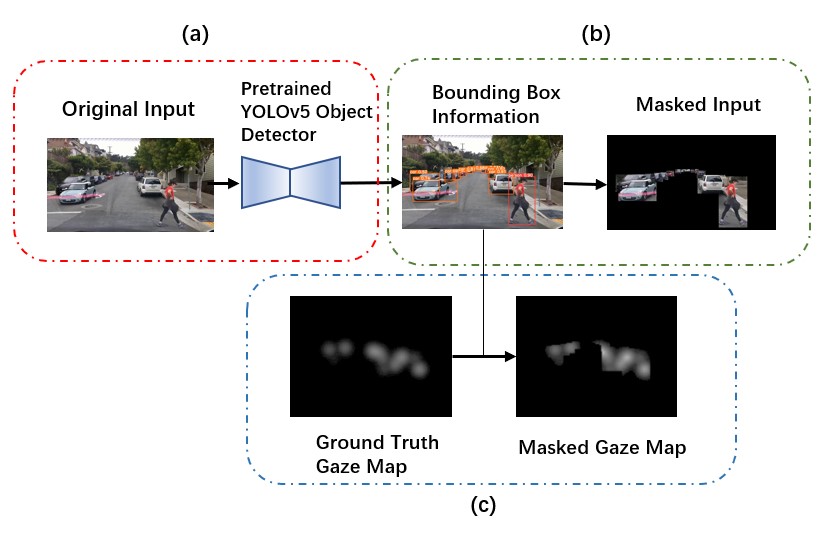}
	\caption{Dataset Cleansing Process.}
	\label{DCP}
\end{figure}

Figure~\ref{DCP} outlines our proposed pipeline for cleansing existing driving gaze datasets. As depicted in Figure \ref{DCP} (a), we first use a pre-trained YOLOv5 \cite{glenn_jocher_2020_4154370} object detector to obtain the bounding box information from the original inputs. 
The detail of this part is given in Section \ref{UG}. Subsequently, using this detection model, we can generate masked inputs (Figure \ref{DCP} (b)) and masked gaze maps (Figure \ref{DCP} (c)) by applying bounding box-specified masks to the original images and gaze maps.
The detail of this part is presented in Section \ref{MG}.


	

\subsection{Unrelated Gaze}
\label{UG}

The driving gaze datasets collected from human drivers contain human bias \cite{ahlstrom2021eye,kotseruba2022attention}. Currently, the major method to collect human attention is to use eye-tracking equipment, therefore drivers' attention can be drawn by unrelated objects, which could hinder the effectiveness of the collected data for training the model. Here, we define the \textit{unrelated gaze} as the collected attention that is not directly related to the driving activity. For instance, if a driver's attention is not focused on an object or 
a 
focused object 
is not related to driving, then this gaze will be regarded as an unrelated gaze.  
By investigating current large-scale driving datasets such as BDD-100K~\cite{bdd100k}, we observe that these datasets only provide annotations for specific objects occurring in driving scenes. For example, BDD-100K \cite{bdd100k} provides object labels for 10 objects for driving-related tasks including 
pedestrian, 
rider, 
car, 
truck, 
bus, 
train, 
motorcycle, 
bicycle, 
traffic light, 
and 
traffic sign. 
It is natural to think that such objects are thought as important objects in driving scenes, 
which 
need to be carefully observed by drivers. Therefore, in this work, we propose to cleanse gaze maps by only keeping gazes on such important objects. To do this,  we fine-tune an object detection model YOLOv5 \cite{glenn_jocher_2020_4154370} and use it to extract the bounding box information of those objects. With the bounding box information, we can remove the unrelated gaze from the original gaze map to obtain the cleaned gaze map (Figure \ref{DCP} (a)).


\subsection{Masked Input and Ground Truth} 
\label{MG}
Several feature extraction approaches, whether based on CNN or VIT \cite{dosovitskiy2010image}, extract features from entire input images without weighing the importance of image segments based on semantic meanings. In gaze prediction, some segments bear greater relevance than others. For instance, objects like vehicles and pedestrians offer crucial driving information compared to less relevant elements like the sky or trees. With prior objects' labels and embedding knowledge, we can consider non-critical objects as noise within the current context.
As a result, we utilize the bounding box information to identify and select critical objects while masking out all other pixels (Figure \ref{DCP} (b)) in our cleansing method.

Apart from the label and embedding information, the spatial location of objects is also crucial in the original inputs. Our cleansing method intends to retain as much location information as possible while removing unnecessary objects. In addition to modifying the input, we also process the ground truth gaze map to eliminate labeling noise (e.g., gaze focused on the sky or trees). To accomplish this, we apply the bounding box technique to the original gaze map, selecting only the gaze information located inside the bounding box for those semantically relevant objects (Figure \ref{DCP} (c)).
\begin{figure*}[h]
    \centering
    \includegraphics[width=1\linewidth]{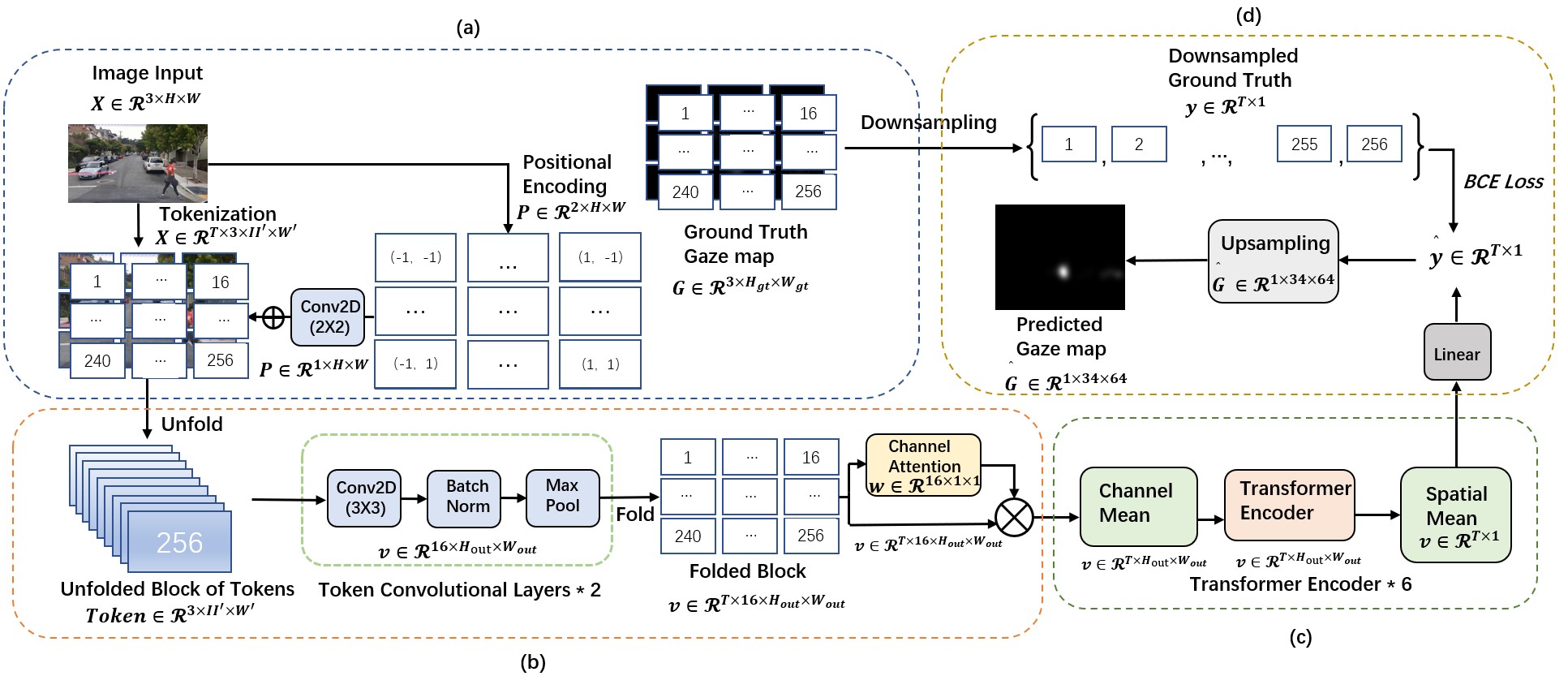}
	\caption{The overall architecture of our proposed CUEING model.}
	\label{Model Structure}
\end{figure*}

 \section{CUEING Model}
 \label{CM}
The overall architecture of our proposed model is shown in Figure \ref{Model Structure}. To design a light-weight gaze prediction model, the first step is to conduct tokenization and positional encoding on original inputs, and then downsample the ground truth gaze map (Figure \ref{Model Structure} (a)). After that, to further extract information from these tokens, we stack up these tokens and apply token convolutional layers (Figure \ref{Model Structure} (b)) and the transformer encoders on them (Figure \ref{Model Structure} (c)). Finally, we calculate the loss and upsample the output of the linear layer to visualize the gaze map (Figure \ref{Model Structure} (d)).




\subsection{Tokenization} 
Grid-based attention prediction yields satisfactory results in the gaze prediction task, as noted in \cite{rong2022and}. However, their approach uses a 1 × 1 kernel size in the convolutional layer to map the extracted features to a gaze value array. This 
does not fully harness the benefits of the ``grid-based'' approach. In our model, we manipulate directly the original image in a ``token-to-point'' approach. We first tokenize original inputs from driving scenarios $X \in \mathcal{R}^{3 \times H \times W}$ to $ X \in \mathcal{R}^{T \times 3 \times H' \times W'}$ where $H$ and $W$ refer to the height and width of original input and $T$ is the number of tokens which is 256 by default. However, it can be adjusted to any power-of-2 value and a perfect square. After conducting the experiments, we determined that 256 strikes a favourable balance between performance and computational efficiency. Since there is the same number of tokens in each row and column, then $H' = H/\sqrt{T}$ and $W' = W/\sqrt{T}$. Specifically, similar to convolution, we first use a sliding window that has the same size as the token ($\mathcal{R}^{3 \times H' \times W'}$) to move through the input from the left-top to right-bottom without any overlap to form tokens. The sliding window is a filter with all parameters as 1 and is not trainable. We call this operation \emph{tokenization}.  Different from convolution, we calculate the dot product between the filter and the token at each movement but with no sum. The process is illustrated in Figure \ref{Model Structure} (a). 

Notably, this operation does not need any trainable parameters and floating-point operations. This contributes to making our model lightweight. Our tokenization is driven by two primary motivations. First, we aim to decrease the Giga Floating Point Operations Per Second (GFLOPs) in the subsequent convolution process. Second, we intend to map a sequence of tokens to a set of points. These points depict the intensity of the gaze likely associated with each respective token. To match the second motivation, we downsample the ground truth gaze map into points using a similar idea with tokenization. Differently, after receiving the tokens from the ground truth gaze map, we calculate an average gaze value (scaled to between 0 and 1) for each token in the ground truth as the `point'. The ground truth gaze map is then downsampled from $ G \in \mathcal{R}^{3 \times H_{gt} \times W_{gt}}$ to $ y \in \mathcal{R}^{ T \times 1}$ with the number of points equal to the number of tokens. The default number of points is 256.  The downsampled ground truth will be used to calculate the loss (Figure \ref{Model Structure} (d)). To get ready for the forthcoming token convolution and attention blocks, we implement the positional encoding for those tokens.
\subsection{Positional Encoding} 
\begin{figure}[h]
    \centering
    \includegraphics[width=0.6\linewidth]{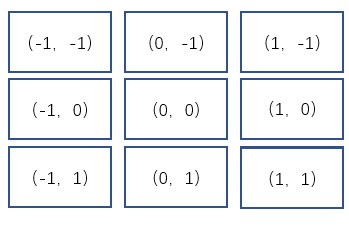}
	\caption{Example of Positional Encoding}
	\label{PE}
\end{figure}

Due to the tokenization and the following token convolution and attention blocks, we assign each token a specific positional in the original input (Figure \ref{Model Structure} (a)). To obtain accurate positional information on tokens, we 
prefer that the location information could be feature-like information in the original input. Therefore the location information can be further processed as a feature by our token convolutional layers in the next stage. Here we employ a specifically designed absolute positional encoding, 
and assign a two-dimensional coordinate to each token.
These coordinates will equally spread between -1 and 1 through the rows and columns. For example, if we tokenize an input into $3 \times 3$ tokens, then the first token has coordinates $(-1,-1)$, while the coordinates to its right will be $(0,-1)$ and the coordinates below it will be $(-1,0)$ (Figure \ref{PE}).
Hence, we can ensure that each token in the original input has a unique coordinate. However, to allow the positional information to be the same dimension as the features of the tokenized input, we use a convolutional layer map all two-dimensional encoded coordinates from $ P \in \mathcal{R}^{ 2 \times H \times W}$ to $ P \in \mathcal{R}^{ 1 \times H \times W}$, and then 
perform 
an element-wise addition between the positional encoding and those tokens as $X = P \oplus X$, where $X$ is the tokenized input.

\subsection{Token Convolution} 
To 
perform 
token convolution, we stack all the tokens from a single input into a batch, which is indicated as the first step of (b) in Figure \ref{Model Structure}. The operation is called \emph{unfold}. Therefore, the dimension in this stage transforms from $ \mathcal{R}^{T \times 3 \times H' \times W'}$ to $ \mathcal{R}^{3 \times H' \times W'}$ with $T$ channels. This is essentially an image-to-column transformation instead of a dimensionality reduction. Then, we use the two convolutional layers with a kernel size of $3 \times 3$ to downsize each token and use 16 kernels
(Figure \ref{Model Structure} (b)), the output feature map for each token is $v \in  \mathcal{R}^{16 \times H_{out} \times W_{out}}$. The optimal number of channels was determined through our experiments. Our findings indicate that utilizing a larger channel size (256 or 512) does not yield significant performance improvements. However, it substantially escalates the demand for computational resources. The advantages of token convolution include reduced GFLOPs. This reduction is achieved as the dimensions of each token, both in terms of width and height, are scaled down by a factor of $\frac{1}{\sqrt{T}}$ from the original input dimensions. According to the GFLOPs required in the convolution operation, the GFLOPs needed for token convolution are merely $\frac{1}{T}$ of those required for standard convolution \cite{deng2019drivers}.


After token convolution, we prepare the stacked tokens for attention blocks. Given our interest in inter-token relationships, we need to perform self-attention between these tokens. We then reshape the convolved token column back to the shape before the unfold operation, and this is the operation we call \emph{fold} in Figure \ref{Model Structure} (b). In this step, the dimension of the feature map $v$ for each token will be laid out flat and the resulting feature map will be $ \mathcal{R}^{T \times 16 \times H_{out} \times W_{out}}$. 

To enable self-attention blocks to process our tokens, we must reduce the dimensionality of the token's feature map. We first merge the height and width dimensions of the token's feature map. To tolerate our interest in inter-token relationships, while also maintaining computational efficiency, we prioritize retaining the merged height and width dimension and processing the channel dimension before the attention block. Here we employ channel attention \cite{woo2018cbam} to the output of the convolutional layer to assign each channel a trainable weight, i.e., $v = v \otimes w$, where $w$ is the weight of channels. By applying weight to each channel, we average the channel dimension, and the dimension of the token's feature is reduced from $ \mathcal{R}^{T \times 16 \times H_{out} \times W_{out}}$ to $ \mathcal{R}^{T \times H_{out} \times W_{out}}$

\subsection{Attention Between Tokens}
Under our settings, the relationship between tokens is important. Ideally, after attention layers, pixels that could generate gaze should become similar. Therefore, we use transformer encoders \cite{vaswani2017attention} for our model. Specifically, we use the symmetric self-attention layer to find the attention scores between tokens, since we want to capture the relationship between all the tokens (Figure \ref{Model Structure} (c)). We validate the importance of attention blocks in our ablation study (Section \ref{Generalizability_result}). 
After the transformer encoders, in order to allow the output 
to 
be processed by linear layer, we use a spatial mean in each token to reduce the dimension, then the dimension is reduced from $ \mathcal{R}^{T \times H_{out} \times W_{out}}$ to $ \mathcal{R}^{T \times 1}$.
We use the output from the linear layer to calculate the loss, and the loss function we use is Binary cross-entropy loss. Equation \ref{loss} indicates the loss function, where $y$ is the downsampled ground truth, $\hat{y}$ is the output of the linear layer.
\begin{equation}
\label{loss}
   L(\hat{y}, y)=-\frac{1}{T} \sum_{i=1}^T y_i \cdot \log \left(\hat{y}_i\right)+\left(1-y_i\right) \cdot\left(1-\log \left(\hat{y}_i\right)\right)
\end{equation}

 Finally, to visualize the gaze map, we use interpolation and Gaussian smoothing to upsample the output from the linear layer to form the final gaze map (Figure \ref{Model Structure} (d)).






\section{Experimental design}
\label{EX}

We proposed four research questions (RQs) and conducted corresponding experiments to evaluate the effectiveness of our proposed model and cleansed datasets: 

\begin{itemize}
\item \textbf{RQ1:} Can our cleansed datasets produce more reasonable human gaze than the baseline dataset?

\item \textbf{RQ2:} To what extent do cleansed datasets enhance the performance of human gaze predictions when compared to using baseline (uncleansed) datasets?

\item \textbf{RQ3:} How does the generalizability of our human attention model compare to existing gaze prediction models across diverse datasets?
\item \textbf{RQ4:} Can our human attention model be efficiently deployed on resource-constrained devices, like mobile devices, without compromising on performance? 

   
    



     
\end{itemize}

    
   
    



     

\subsection{Experimental Setup}
To evaluate those research questions, based on two large and popular driving gaze datasets BDD-A \cite{xia2018predicting} and DADA-2000 \cite{fang2019dada}, we produced two cleansed datasets CUEING-B and CUEING-D for comparison. We resized all inputs into  $1280 \times 720$ which is aligned with the initial size of BDD-A \cite{xia2018predicting} input. In baseline model selection, we included the CUEING model and the latest and popular driving gaze prediction models, including HWS \cite{xia2018predicting}, CDNN \cite{deng2019drivers}, and Where\&What \cite{rong2022and}. Our code implementation is based on Pytorch \cite{NEURIPS2019_9015} and used the Adam optimizer \cite{kingma2014adam}. We conducted experiments for RQ1, RQ2 and RQ3 on an RTX 3090 GPU, while RQ4 experiments were performed on a Jetson Nano 4G.

\subsection{Experimental Settings}





\subsubsection{RQ1: User Study} In order to qualitatively understand if our cleansed datasets produce more reasonable human gaze maps than the baseline dataset, we conducted a user study to compare the BDD-A \cite{xia2018predicting} with the CUEING-B and the DADA-2000 \cite{fang2019dada} with the CUEING-D. BDD-A \cite{xia2018predicting} and CUEING-B contain 30,073 images, and DADA-2000 \cite{fang2019dada} and CUEING-D contains 22,171 images. To ensure a 95\% confidence level with a 10\% confidence interval, we randomly sampled 100 images with gaze maps from both original datasets and their cleansed version, respectively. In those sampled groups, to gain quantitative measurement, we let users decide how many clusters of gaze each image contains and how many of them are reasonable. For example, 
in Figure 4, 
the left image has 4 clusters of gaze, and 1 of them is unreasonable, and the right image has 2 clusters of gaze, and all of them are reasonable. We recruited 10 workers in Amazon mTurk \cite{crowston2012amazon} to participate in the user study, and all of them are car owners and have extensive driving experience. They can receive \$1 for each survey they take. We made 5 surveys in total.  During the user study, we did not instruct users on what kinds of gazes are reasonable. Users need to rely entirely on their subjective perceptions to determine which image contains a more reasonable gaze.

After collecting human study results, we calculated statistical information including the mean values of the total number of gazes and number of reasonable gazes, and the ratio of reasonable gazes over the total number of gazes on all datasets.  

\begin{figure}[!t]    
    \centering
    \includegraphics[width=1\linewidth]{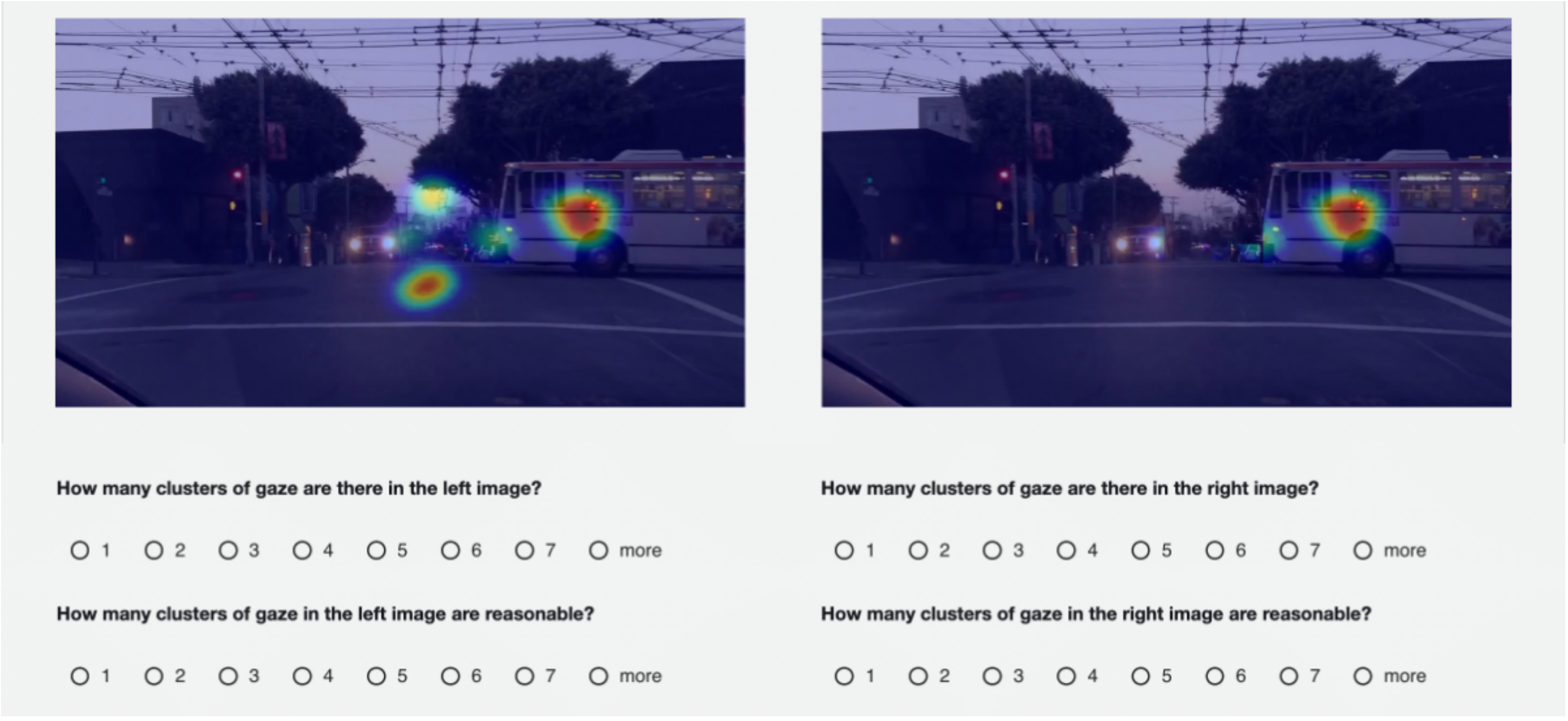}
	\caption{Example of user study interface, the left side is the overlay of image and gaze map from DADA-2000, the right side is the overlay of image and gaze map from CUEING-D. }
\label{us}  
\end{figure}

\subsubsection{RQ2: Cleansing Method} 
\label{sec:exp_rq2}
To evaluate the effectiveness of the proposed cleansing method, 
following the principle ``training on synthetic and testing on real'' \cite{varol2017learning, peng2018visda}. We trained the gaze prediction models on the original dataset and cleansed (synthesized) dataset respectively, but the evaluations of the models are only on the original datasets. If the performance of gaze prediction models increases, the cleansing method is proven effective in enhancing gaze prediction models. We chose DADA-2000 \cite{fang2019dada} as our original dataset. The ground truth of DADA-2000 \cite{fang2019dada} is sourced from a single driver, the unrelated gaze during driving is more noticeable when compared to the averaged gaze collected from multiple drivers.

\par The performance of gaze prediction models on different datasets was measured from the object level. 
We measured whether the gaze prediction models can focus the gaze on the correct objects in the ground truth gaze map, which is measured by metrics Accuracy, Precision, Recall, F1, and AUC. 
In order to determine whether an object is considered `focused', we defined a threshold criterion based on the gaze predictions within the bounding box of the detected object. Following the approach used in \cite{rong2022and}, if the maximum predicted gaze value within the bounding box exceeds the threshold of 0.5 (from the empirical study of \cite{rong2022and}), the object is deemed as being focused on by the driver. This threshold is independent of the training and pixel-level metrics and is used primarily for object-level evaluation. 
\par We also compared the proposed cleansing method with another data processing method SAGE~\cite{pal2020looking}. We used SAGE to create datasets SAGE-B from BDD-A and then followed the same process to train gaze prediction models on the synthesized datasets. We compared the performance of trained models on SAGE-B with models trained on CUEING-B to identify which method is more effective in improving the performance of gaze prediction models. The evaluation metrics used the same as above.

\subsubsection{RQ3: Generalizability} 
We assessed the model's generalizability using two methods. First, a model has strong generalizability if it has advanced performance on the target dataset directly after training on the source dataset. Second, a model was deemed to be more generalizable if, after being trained on the source dataset, its performance was improved when fine-tuned with just 2\% of the target dataset's training data. 

In the experiment, to avoid the possible influence of the proposed dataset cleansing method, we only employed the original BDD-A \cite{xia2018predicting} as the source datasets and used DADA-2000 \cite{fang2019dada} and Dr(eye)VE \cite{alletto2016dr} as target datasets respectively. To evaluate the second kind of generalizability, we randomly sampled 2\% of images from the corresponding target datasets for fine-tuning gaze prediction models.

We used the object level metrics introduced in Section~\ref{sec:exp_rq2} to assess the models' general performance and generalizability in this RQ. Meanwhile, we introduce additional pixel level metrics for model evaluation. For the pixel level, we measured the similarity between the generated gaze maps and the ground truth gaze maps, using Kullback–Leibler divergence ($D_{KL}$) and Pearson’s Correlation Coefficient ($CC$) metrics as in previous works \cite{xia2018predicting, rong2022and}. In RQ2, we did not evaluate our cleansing method at the pixel level. Given our approach of "training on synthetic and testing on real", it's understood that synthesized data differs in distribution from real data.


\subsubsection{RQ4: Mobile Device Testing}
To assess if the model is suitable for deploying on a mobile device, we first compared different models' complexity. We employed the same metrics as in \cite{rong2022and}, which include the number of trainable parameters and GFLOPs. These metrics provide a standardized measure for assessing the complexity of different models.

\begin{figure}[!t]    
    \centering
    \includegraphics[width=0.6\linewidth]{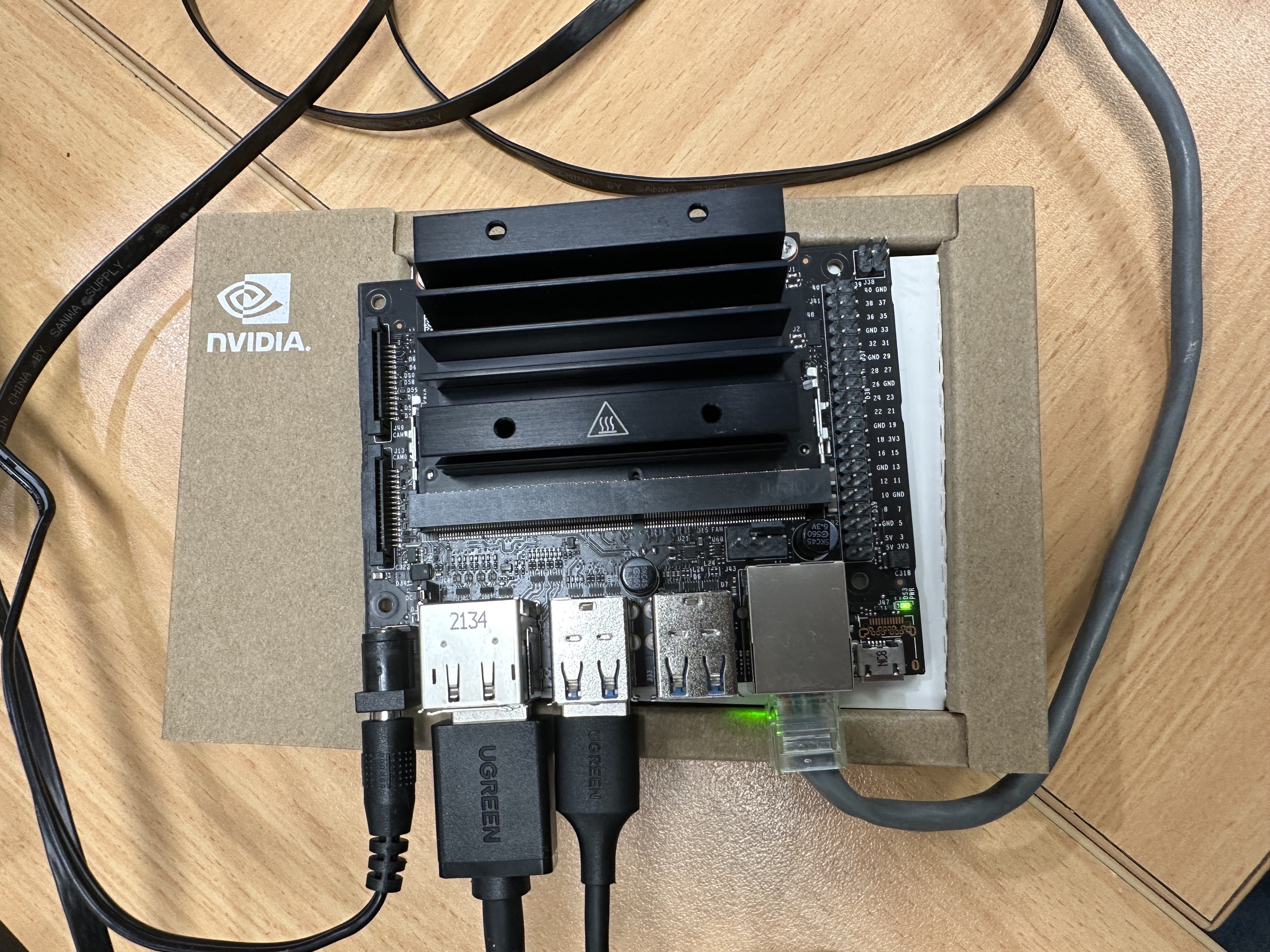}
	\caption{The Jetson nano we used to conduct the mobile device experiment.}
\label{nano}  
\end{figure}

We assessed the efficiency of the two gaze prediction models CDNN \cite{deng2019drivers} and CUEING on a mobile device with Nvidia Jetson Nano 4G (Figure \ref{nano}). We selected CDNN as the comparison baseline because, among the baseline models, it has the second lowest model complexity. Additionally, while CDNN uses standard CNN for input processing, our method utilizes token convolution. On the mobile device, we assessed the energy consumption, processing unit overhead, and inference speed for both models on processing a single image as these metrics are crucial for model deployment on mobile devices.

\section{Results}
\label{Result}
\subsection{RQ1: User Study}

\begin{figure}[!t]    
    \centering
    \includegraphics[width=0.8\linewidth]{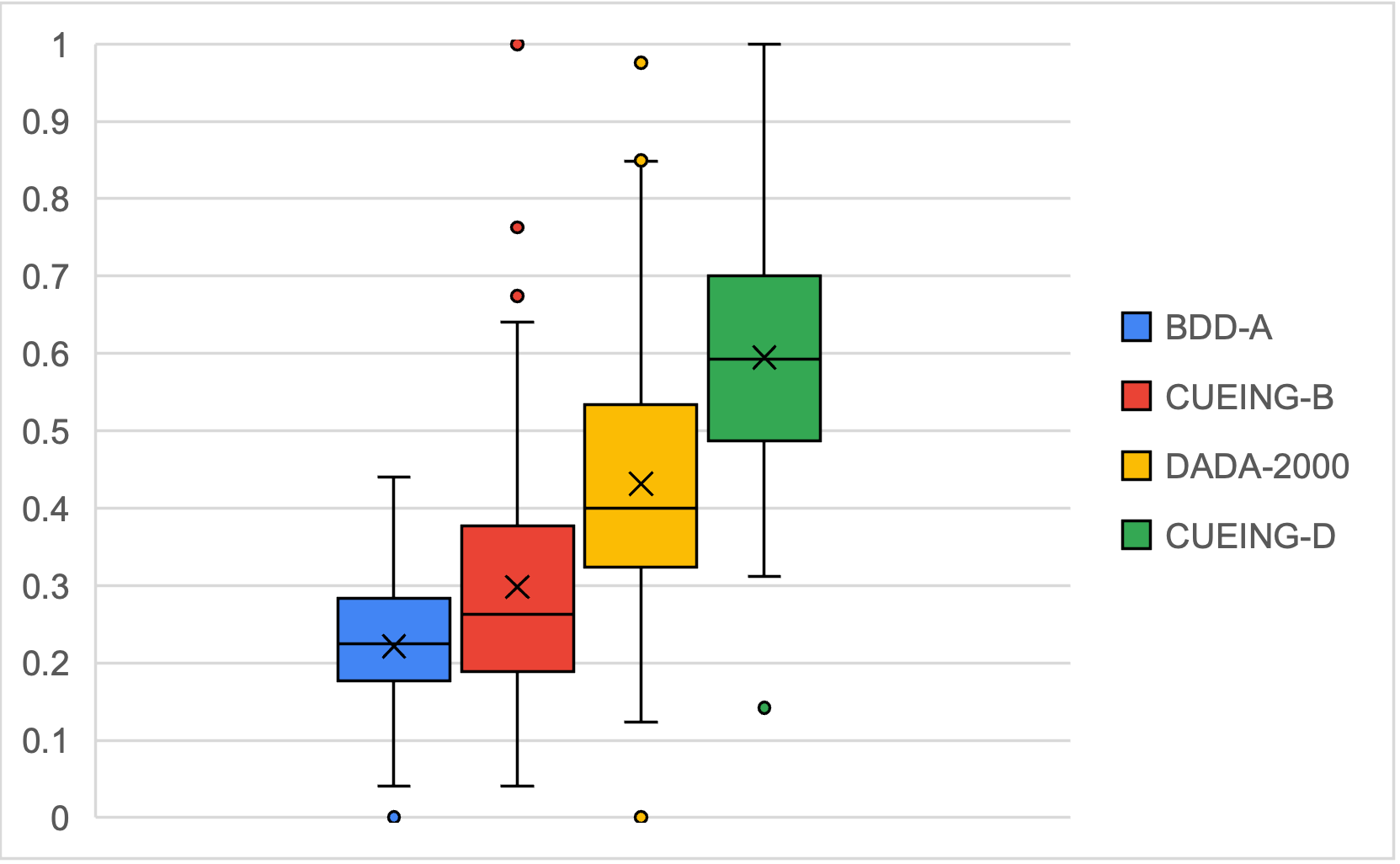}

	\caption{Quantitative result of the user study. The order of the boxes in the figure is BDD-A, CUEING-B, DADA-2000, and CUEING-D (left to right).}
\label{usr}  
\end{figure}

To process our collected user study results from mTurk \cite{crowston2012amazon}, we first calculated the mean value using the answers from all users for each question. We then used these mean values to find the ratio of reasonable gaze in each image, and finally, we Min-Max normalized the ratio from the original dataset and its cleansed version to obtain the quantitative results. Figure \ref{usr} indicates the quantitative result of our user study. We can see that users consider the datasets that have used our cleansing method (red and green) to contain a more reasonable gaze than their corresponding original datasets (blue and yellow) in both medians and means.

\begin{figure}[!t]    
    \centering
    \includegraphics[width=0.8\linewidth]{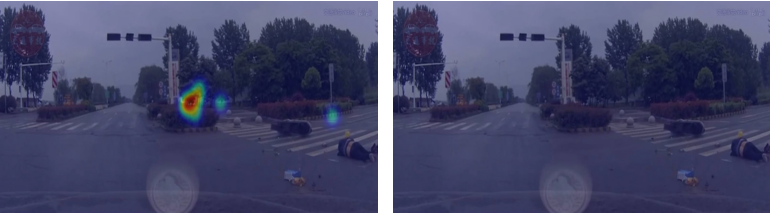}
    \includegraphics[width=0.8\linewidth]{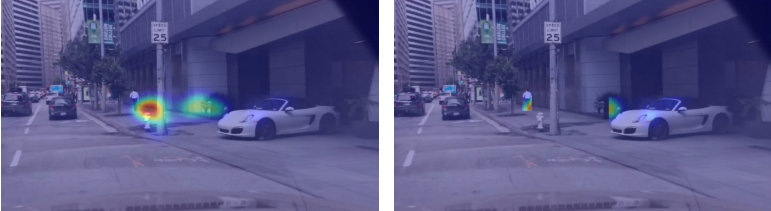}
	\caption{Outlier examples, the first row is from DADA-2000 (left) and CUEING-D (right), the second row is from BDD-A (left) and CUEING-B (right). }
\label{lo}  
\end{figure}

Here, we further explore the reasons for the outliers in Figure \ref{usr}. In Figure \ref{lo}, we present some examples of outliers. The initial row showcases examples from DADA-2000 (on the left) and CUEING-D (on the right). After our cleansing process, our sample has no gaze. This highlights a limitation in our cleansing approach. Given that we train the YOLOv5 model \cite{glenn_jocher_2020_4154370} using labels from BDD-100K \cite{bdd100k}, the detectable objects are confined by the label availability in BDD-100K.
In this instance, it is reasonable for the driver to look at the bushes in this circumstance, but the bush is not in the BDD-100K's \cite{bdd100k} label set. 
This points out an interesting future research 
direction 
where we can generate more valuable label sets for existing driving gaze datasets.

The second row in Figure \ref{lo} displays examples from BDD-A \cite{xia2018predicting} and CUEING-B. In the second row, the left image from BDD-A \cite{xia2018predicting} demonstrates a scenario where both datasets fail to detect the white car, despite it being the object of focus for most drivers in similar situations. However, 
it is 
important to note that this image represents a single frame from the video, and it is possible that the driver may have shifted their attention to the car later in the video. This highlights the significance of incorporating time series data in gaze prediction rather than relying solely on individual frames. Previous research has indicated that integrating time series data at this stage does not enhance performance and may introduce significant centre bias \cite{rong2022and}. As a potential future direction, training a robust gaze attention model on videos is essential.


\subsection{RQ2: Cleansing Method} 
Table \ref{general performance DADA} and \ref{general performance} list the evaluation result of RQ2.  
In Table \ref{general performance DADA}, most models trained by our cleansed datasets have different levels of improvement in object-level metrics. 
Specifically, the CUEING model and Where\&What \cite{rong2022and} are improved up to 7.38\% and 6.25\% in accuracy, 8.75\% and 3.57\% in AUC.


\begin{table*}[!htbp]
\caption{Selected models' performance trained by CUEING-D and DADA-2000 datasets, and test on DADA-2000 test set. The best result is bold.}

\centering
\scalebox{1}{
\begin{tabular}{l|lllll}
\hline
\multicolumn{1}{c|}{Model}& \multicolumn{5}{|c}{Object Level (\textbf{CUEING-D}/DADA-2000)} \\
\cline{2-6}
\multicolumn{1}{c|}{}& Acc($\%$) & Prec($\%$) & Recall($\%$) & F1($\%$) & AUC \\
\hline

{CDNN}&41.18/\textbf{47.04} & 25.89/\textbf{25.94} & \textbf{84.74}/71.45 & \textbf{39.66}/38.06 & 0.58/\textbf{0.59}    \\
{Where\&What}&\textbf{85.14}/78.89   &46.81/\textbf{63.14}  &\textbf{65.71}/65.54 &54.67/\textbf{64.32} &\textbf{0.87}/0.84   \\
{HWS}&\textbf{47.16}/45.79   &25.04/\textbf{25.43} &66.04/\textbf{71.45}   &36.32/\textbf{37.51} 
&0.55/\textbf{0.57} \\
{\textbf{CUEING}}& \textbf{85.80}/78.42  &48.42/\textbf{52.50} & \textbf{62.12}/57.04 & 54.42/\textbf{54.68}  &\textbf{0.87}/0.80  \\
\hline
\end{tabular}
}

\label{general performance DADA}
\end{table*}

\begin{figure}[!tb]
\centering
\begin{subfigure}[b]{0.2\textwidth}
        \includegraphics[width=\textwidth]{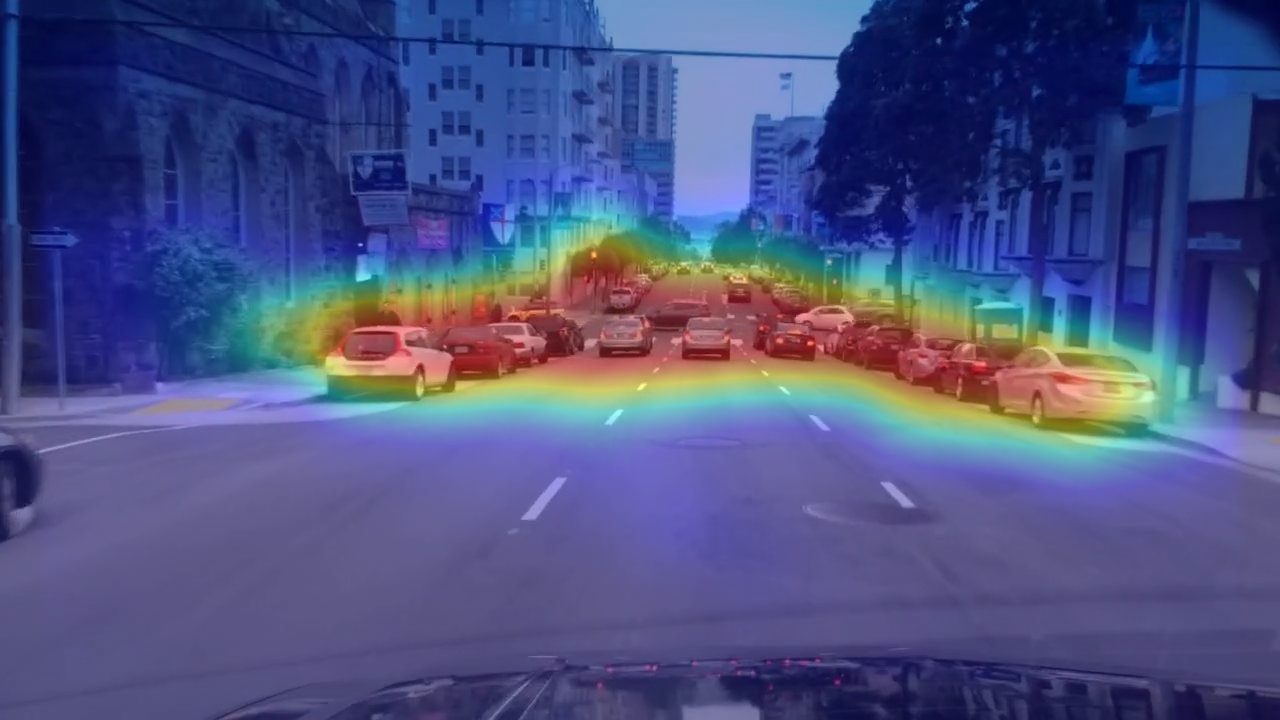}
        \includegraphics[width=\textwidth]{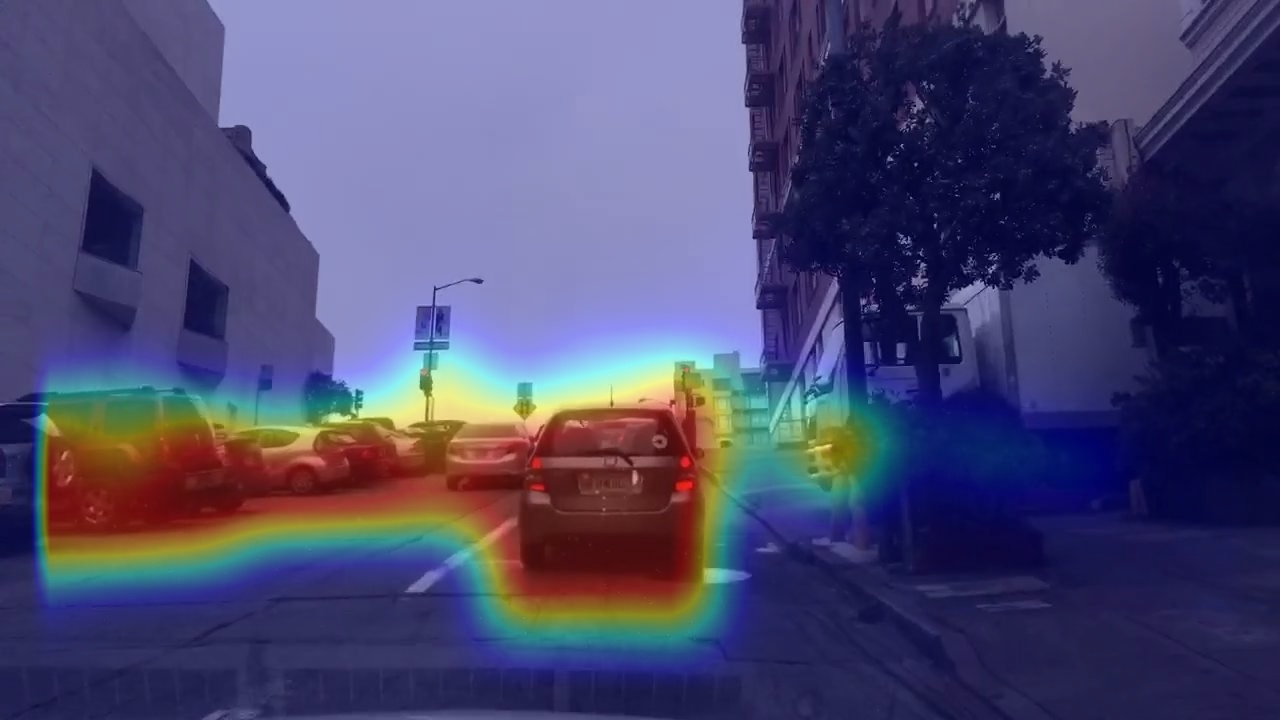}
        \caption{SAGE-B}
        \label{SAGE-B}
\end{subfigure}
\begin{subfigure}[b]{0.2\textwidth}
        \includegraphics[width=\textwidth]{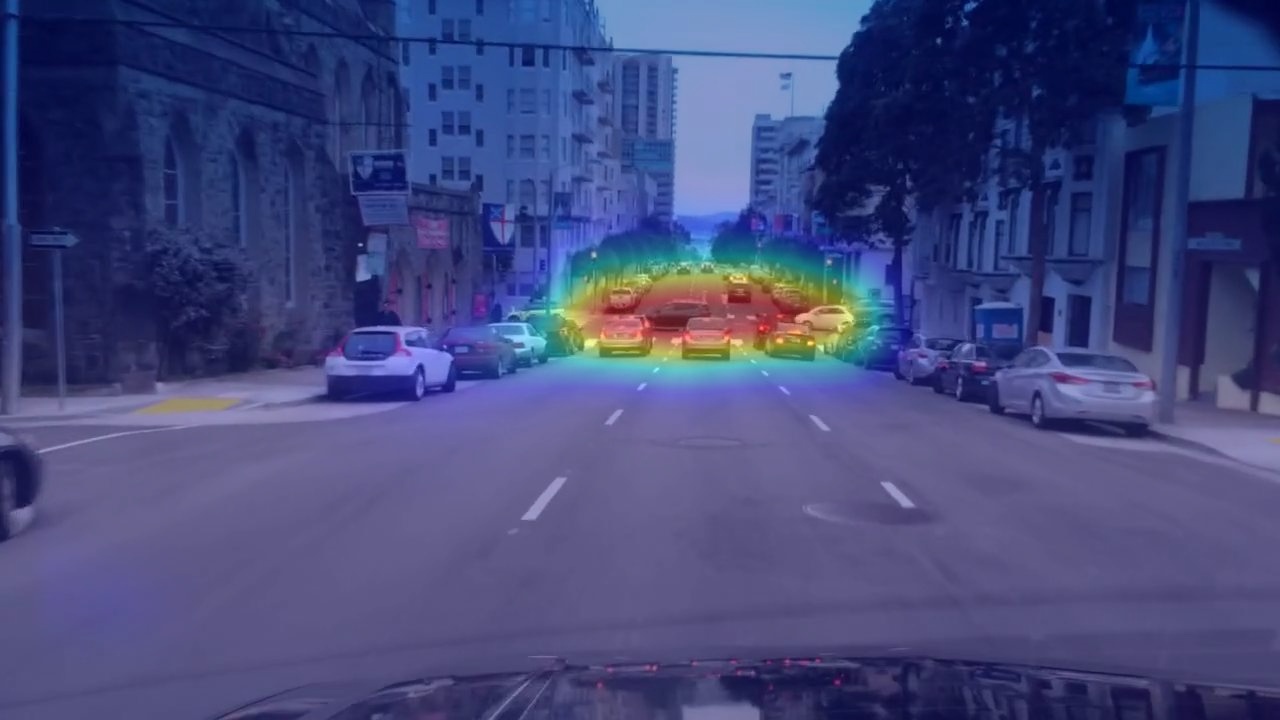}
        \includegraphics[width=\textwidth]{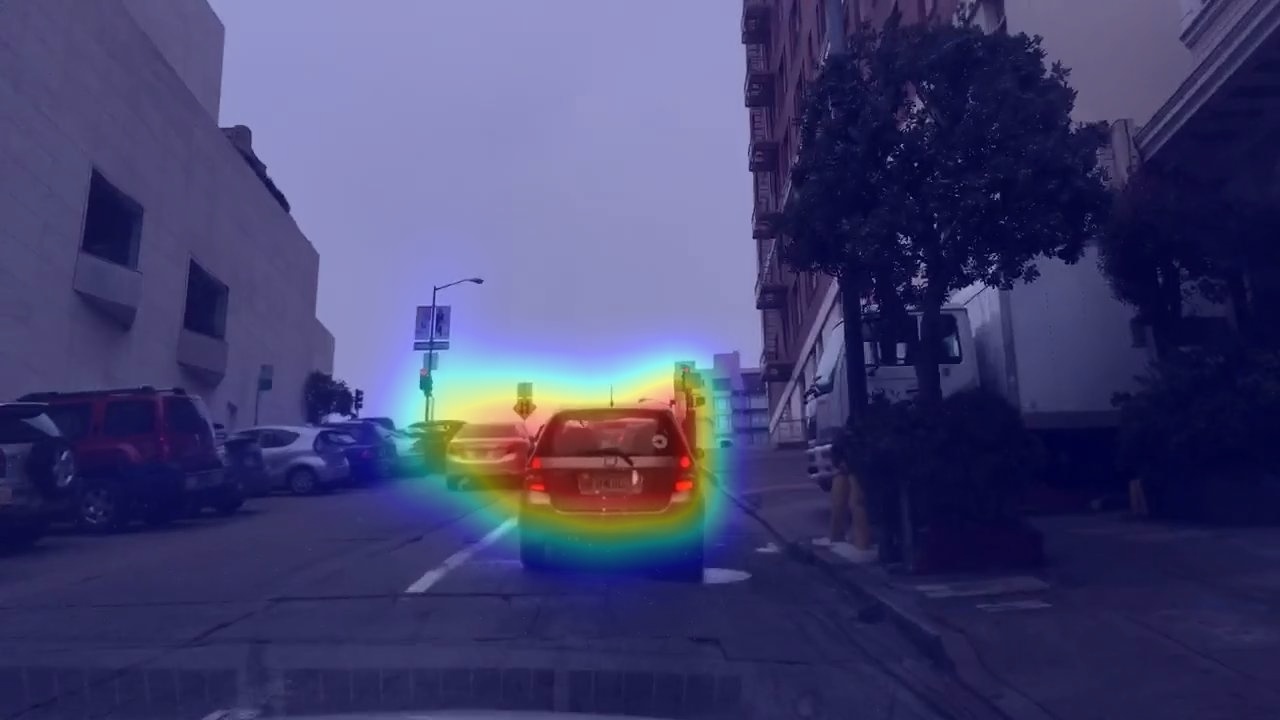}
        \caption{CUEING-B}
        \label{SAGE-B}
\end{subfigure}

\caption{Example of prediction results generated by the CUEING model trained on SAGE-B (left) and CUEING-B (right).}

\label{SVSC}
\end{figure}

In Table \ref{general performance}, under the principle of `training on synthetic and testing on real', the performance of SAGE-B \cite{pal2020looking} is significantly worse than the performance of CUEING-B, but it is worth noting that the recalls of the models trained on SAGE-B \cite{pal2020looking} are very high.  This is because SAGE \cite{pal2020looking} almost includes all objects' semantic segmentation from the input in their ground truth. Figure \ref{SVSC} indicates a comparison between the generated gaze map using the CUEING model trained by SAGE-B \cite{pal2020looking} and CUEING-B, the gaze map generated by the model trained by SAGE-B \cite{pal2020looking} is quite different from real human's gaze. Meanwhile, SAGE \cite{pal2020looking} tends to capture all objects in the input instead of focusing on the important objects, which hinders the ability of models to capture important objects to benefit ADS's decision-making. As a future direction,  we intend to evaluate the importance of objects identified by the gaze model when co-training critical driving models. The goal is to enhance the robustness of self-driving models, making them aligned with human knowledge.

\begin{table*}[!htbp]
\caption{Selected models' performance trained by CUEING-B and SAGE-B datasets, and test on BDD-A test set. The best result is bold.}

\centering
\scalebox{1}{
\begin{tabular}{l|lllll}
\hline
\multicolumn{1}{c|}{Model}& \multicolumn{5}{|c}{Object Level (\textbf{CUEING-B}/SAGE-B)}  \\
\cline{2-6}
\multicolumn{1}{c|}{}& Acc($\%$) & Prec($\%$) & Recall($\%$) & F1($\%$) & AUC  \\
\hline

{CDNN}& \textbf{53.48}/38.76  & \textbf{44.27}/37.74  & 97.26/\textbf{99.78}  & \textbf{60.85}/54.77  & \textbf{0.84}/0.60      \\
{Where\&What}&\textbf{78.11}/53.42   &\textbf{67.60}/44.93 &72.19/\textbf{90.46}  &\textbf{69.70}/60.05&\textbf{0.84}/0.63    \\

{HWS}&\textbf{57.19}/49.65   &\textbf{46.21}/41.91 &\textbf{92.54}/91.92   &\textbf{61.64}/57.57
&\textbf{0.78}/0.62  \\
{\textbf{CUEING}}& \textbf{77.39}/59.10  & \textbf{71.64}/47.17  & 68.84/\textbf{81.86} & \textbf{70.21}/59.44  & \textbf{0.84}/0.68 \\
\hline
\end{tabular}
}
\label{general performance}
\end{table*}

\subsection{RQ3: Generalizability} 
\label{Generalizability_result}


In Tables \ref{ft-dada} and \ref{ft-dreye}, we present the evaluation of models' generalizability on two target datasets.
Our CUEING model consistently outperforms the other baseline models, both with and without fine-tuning. For example, in Table \ref{ft-dada}, without fine-tuning, the accuracy of CUEING reaches 58.67\%, which outperforms the current state-of-the-art (e.g., Where\&What \cite{rong2022and}) by 3.54\%. When fine-tuning with a portion of DADA-2000 data \cite{fang2019dada}, the CUEING model surpasses the state-of-the-art (e.g., Where\&What \cite{rong2022and}) by up to 12.13\% in the pixel level metrics (KL value of 2.10 versus 2.39).

\par In Table \ref{ft-dreye}, similar outcomes are evident in the generalization task for Dr(eye)VE \cite{alletto2016dr}. The CUEING model exhibits a notable advantage in precision, achieving 67.99\% without fine-tuning, and its performance in other metrics closely approaches the state-of-the-art.


\begin{table*}[!t]
\caption{Model performance by using BDD-A as source dataset Fine-tuned with DADA-2000 as target dataset.  The best result is bold. In the table,
* indicates that the model is fine-tuned on the corresponding dataset, and † informs the direct prediction without fine-tuning.}
\centering
\scalebox{0.9}{
\begin{tabular}{l|lllll|ll}
\multicolumn{8}{c}{}\\
\hline
\hline
\multicolumn{1}{c|}{Model}& \multicolumn{5}{|c|}{Object Level} &\multicolumn{2}{c}{Pixel Level} \\
\cline{2-8}
\multicolumn{1}{c|}{}& Acc($\%$) & Prec($\%$) & Recall($\%$) & F1($\%$) & AUC & KL & CC \\
\hline

{CDNN*}&36.93    &24.84  &\textbf{87.37}  &38.68   &0.53  &3.52  &0.16     \\
{Where\&What*}&\textbf{79.51}  &\textbf{55.49}  &51.58   &53.46 &0.79   &2.35   &0.33    \\
{HWS*}&42.87  &25.19 &76.36 &37.89   &0.56 &5.71  & 0.08 \\

{\textbf{CUEING}*}&78.71  &52.98  &59.55  &56.08  &\textbf{0.80}  &\textbf{2.19} &\textbf{0.35} \\
\hline
{CDNN†}&36.80 &24.84 &\textbf{87.65}  & \textbf{38.71} &0.53  &\textbf{3.15}  &0.05    \\
{Where\&What†}&55.13 &24.42 &46.16  &31.94 &\textbf{0.56}  &3.36  &0.09\\
{HWS†}&46.29 &\textbf{25.34} &69.55  &37.14  &0.55 &4.62 &\textbf{0.13} \\
{\textbf{CUEING}†}&\textbf{58.67} &24.73  &39.70  &30.48  &\textbf{0.56}  &3.65 &0.07  \\
\hline
\end{tabular}
}

\label{ft-dada}
\end{table*}

\begin{table*}[!t]
\caption{Model performance by using BDD-A as source datasets Fine-tuned by Dr(eye)VE as target dataset.  The best result is bold. In the table, 
* indicates that the model is fine-tuned on the corresponding dataset, while † informs the direct prediction without fine-tuning.}
\centering
\scalebox{0.9}{
\begin{tabular}{l|lllll|ll}
\multicolumn{8}{c}{}\\
\hline
\hline
\multicolumn{1}{c|}{Model}& \multicolumn{5}{|c|}{Object Level} &\multicolumn{2}{c}{Pixel Level} \\
\cline{2-8}
\multicolumn{1}{c|}{}& Acc($\%$) & Prec($\%$) & Recall($\%$) & F1($\%$) & AUC & KL & CC \\
\hline
{CDNN*}& 52.17   &45.92   &\textbf{98.72}  &62.68 &0.86  &2.07 &0.40 \\

{Where\&What*}&\textbf{79.45}  & \textbf{73.87}  &76.87 &\textbf{75.34}& \textbf{0.87}&\textbf{1.54 } &\textbf{0.57}  \\


{HWS*}&65.91   &54.96 &89.90 &68.21   &0.83   &3.30   &0.43 \\

{\textbf{CUEING}*}&78.41  &73.19   &74.38  &73.78   &0.85   &1.71 & 0.53  \\
\hline
{CDNN†}&55.45 &47.69  &\textbf{98.20}  &64.20  &\textbf{0.86}  &1.96  &0.44    \\

{Where\&What†}&\textbf{76.54}&67.98  &80.46  &\textbf{73.70}  &0.85  &\textbf{1.76}  &\textbf{0.50}   \\


{HWS†}&67.09  &56.11 &87.77  &68.45  &0.82  &2.96  &0.46\\

{\textbf{CUEING}†}&75.85  &\textbf{67.99}  &77.24  &72.32  &0.83 &1.88  &0.48  \\
\hline
\end{tabular}
}

\label{ft-dreye}
\end{table*}

\textbf{Generalizability Ablation Study.} 
The CUEING model demonstrates good generalizability. However, its workings remain opaque. To determine which component of the model contributes most to its generalizability, we have undertaken an ablation study focused on this aspect.
We compared the results from freezing all layers before the last linear layer and freezing all attention blocks in the CUEING model trained by CUEING-B.


\begin{table}[]
\caption{Ablation Study on Generalizability of CUEING Model. In the table, 
         † indicates that the model is trained via BDD-A, * informs it is trained via CUEING-B Dataset.}
    \centering
    \scalebox{1}{
    \begin{tabular}{lcc}
\hline Freeze Block & KL & CC\\
\hline 
All except Linear* & 2.49&	0.29	  \\
Attention Module*& 2.35&	0.32	 \\
None of the layers*&2.10&	0.38	\\
\hline 
All except Linear† & 2.28&	0.34	  \\
Attention Module†& 2.23&	0.35	  \\
None of the layers†&2.19	&0.36	\\
\hline
\end{tabular}

}

\label{Generalizability_Study}
\end{table}



From Table \ref{Generalizability_Study}, we can find that the performance decreased for the model trained on the CUEING-B dataset is larger than the model trained on BDD-A \cite{xia2018predicting} under all the settings. This is because when we generalized our model to DADA-2000 \cite{fang2019dada}, we did not apply masks on the input of DADA-2000 \cite{fang2019dada}, and the distribution of the input data is different from the CUEING-B dataset largely. Therefore, from the result of the CUEING model trained on the BDD-A \cite{xia2018predicting} dataset, we find that if we transfer our model under a similar distribution, only allowing the last linear to be trained can reach a similar result with training all the parameters, which only has a 2.67\% increase on the KL-divergence. From the result of the CUEING model trained on the CUEING-B dataset, we can find that even if we allow all the blocks other than attention blocks for training, the KL-divergence still increases by 11.11\% than allowing all the parameters to be trained. Clearly, attention blocks are a crucial component of our model.

\subsection{RQ4: Mobile Device Testing}

Table \ref{PA} presents the evaluation metrics for the model complexity by all baseline models. Remarkably, our model delivers commendable results using merely 1.8\% of the model parameters needed by Where\&What \cite{rong2022and}, and the GFLOPs required are just 6.12\% of what Where\&What consumes. Our model distinctly excels in both complexity and performance compared to other baseline models, making it a promising candidate for deployment on vehicles. 

\begin{table}[]
\caption{Model complexity of different gaze prediction models, and the best
results are bold.}
    \centering
    \scalebox{1}{
    \begin{tabular}{lcc}
\hline Model & \textit{Param.(M)} & \textit{GFLOPs}\\
\hline 
CDNN & 0.68&	5.06	  \\
Where\&What& 7.52 &	17.0	 \\
HWS&3.75 &	21.18	\\
\textbf{CUEING} &\textbf{0.14}&\textbf{0.31}\\
\hline 

\end{tabular}

}

\label{PA}
\end{table}

We evaluated CDNN \cite{deng2019drivers} and CUEING on a Jetson Nano 4G. While CUEING consumed 3W for a single image ($1280 \times 720$) inference and took $0.876$s, CDNN \cite{deng2019drivers} was inoperable on the Jetson Nano even with the half-sized input due to its hefty computational requirements. We pruned CDNN \cite{deng2019drivers} using PyTorch \cite{NEURIPS2019_9015}, and it remained too large for the Jetson Nano 4G. We further explored the reason on an Intel 12600-KF CPU, CDNN \cite{deng2019drivers} required $241$Mb of memory and $0.728$s to infer a single image, whereas in the same setting CUEING only required $164$Mb and $0.048$s. Even when halving the input size for CDNN \cite{deng2019drivers}, it still demanded $0.211$s. The processor's ability to process batches in parallel during our tokenization and unfold operations contributes to enhanced performance. While model compression has its merits, the superiority of token convolution in single-image inference highlights the criticality of reducing model computation.




\section{Threat to Validity}
\label{TV}

The external validity is regarding the generalizability of the proposed method. To counter this, we conducted experiments on large-scale datasets including BDD-A \cite{xia2018predicting}, DADA-2000 \cite{fang2019dada}, SAGE \cite{pal2020looking}, and Dr(eye)VE \cite{alletto2016dr}. The experimental results highlighted the effectiveness of the proposed method across different datasets. The construct validity is about the human-centric study in this work. To ensure that the human study can correct reflect the accuracy of gaze maps produced by different methods, we targeted participants who have driving backgrounds and have the real-world expertise to emulate genuine reactions in driving scenarios. The internal validity arises at the choice of YOLOv5 \cite{glenn_jocher_2020_4154370} for the cleansing process. Though various object detection models have been proposed in recent years, YOLOv5 is more lightweight and can achieve the highest running speed while maintaining similar object detection performance~\cite{yolofast} as more advanced object detection models like  YOLOv7~\cite{yolov7}. 

\section{Conclusion}

This paper presents a novel method to address data labeling noise, generalization gaps, and model complexity in driving gaze datasets and models using an adaptive dataset cleansing approach and a lightweight convolutional self-attention gaze prediction model. Our tests validate the effectiveness and efficiency of our model, especially on mobile devices.

Additionally, our work suggests future research directions, such as developing robust gaze models through multi-camera video data and unconventional semantic objects in driving scenes. Further examination of a streamlined gaze prediction model, along with a refined gaze dataset on driving models, could provide valuable insights, potentially enabling online training in self-driving cars to enhance robustness and personalized driving experiences.



\bibliographystyle{IEEEtran}

{\footnotesize
\bibliography{sample-base}

\begin{thebibliography}{10}
\providecommand{\url}[1]{#1}
\csname url@samestyle\endcsname
\providecommand{\newblock}{\relax}
\providecommand{\bibinfo}[2]{#2}
\providecommand{\BIBentrySTDinterwordspacing}{\spaceskip=0pt\relax}
\providecommand{\BIBentryALTinterwordstretchfactor}{4}
\providecommand{\BIBentryALTinterwordspacing}{\spaceskip=\fontdimen2\font plus
\BIBentryALTinterwordstretchfactor\fontdimen3\font minus \fontdimen4\font\relax}
\providecommand{\BIBforeignlanguage}[2]{{%
\expandafter\ifx\csname l@#1\endcsname\relax
\typeout{** WARNING: IEEEtran.bst: No hyphenation pattern has been}%
\typeout{** loaded for the language `#1'. Using the pattern for}%
\typeout{** the default language instead.}%
\else
\language=\csname l@#1\endcsname
\fi
#2}}
\providecommand{\BIBdecl}{\relax}
\BIBdecl

\bibitem{accidentnews}
\BIBentryALTinterwordspacing
O.~Wagner, ``Nearly 400 car crashes in 11 months involved automated tech, companies tell regulators,'' \emph{National Public Radio}, 2022-06-15. [Online]. Available: \url{https://urlzs.com/jGdAF}
\BIBentrySTDinterwordspacing

\bibitem{news2}
\BIBentryALTinterwordspacing
A.~Paul. (2021). [Online]. Available: \url{https://urlzs.com/ewbg7}
\BIBentrySTDinterwordspacing

\bibitem{codevilla2018end}
F.~Codevilla, M.~M{\"u}ller, A.~L{\'o}pez, V.~Koltun, and A.~Dosovitskiy, ``End-to-end driving via conditional imitation learning,'' in \emph{2018 IEEE International Conference on Robotics and Automation (ICRA)}.\hskip 1em plus 0.5em minus 0.4em\relax IEEE, 2018, pp. 4693--4700.

\bibitem{xia2020periphery}
Y.~Xia, J.~Kim, J.~Canny, K.~Zipser, T.~Canas-Bajo, and D.~Whitney, ``Periphery-fovea multi-resolution driving model guided by human attention,'' in \emph{Proceedings of the IEEE/CVF Winter Conference on Applications of Computer Vision}, 2020, pp. 1767--1775.

\bibitem{zhang2021recent}
R.~Zhang, F.~Torabi, G.~Warnell, and P.~Stone, ``Recent advances in leveraging human guidance for sequential decision-making tasks,'' \emph{Autonomous Agents and Multi-Agent Systems}, vol.~35, no.~2, p.~31, 2021.

\bibitem{bao2021drive}
W.~Bao, Q.~Yu, and Y.~Kong, ``Drive: Deep reinforced accident anticipation with visual explanation,'' in \emph{Proceedings of the IEEE/CVF International Conference on Computer Vision}, 2021, pp. 7619--7628.

\bibitem{wang2023decision}
Y.~Wang, J.~Jiang, S.~Li, R.~Li, S.~Xu, J.~Wang, and K.~Li, ``Decision-making driven by driver intelligence and environment reasoning for high-level autonomous vehicles: A survey,'' \emph{IEEE Transactions on Intelligent Transportation Systems}, 2023.

\bibitem{news3}
\BIBentryALTinterwordspacing
P.~Larsson. (2022). [Online]. Available: \url{https://urlzs.com/FZuu9}
\BIBentrySTDinterwordspacing

\bibitem{news4}
\BIBentryALTinterwordspacing
K.~Wiggers. (2020). [Online]. Available: \url{https://urlzs.com/jC5YP}
\BIBentrySTDinterwordspacing

\bibitem{news5}
\BIBentryALTinterwordspacing
A.~James. (2022). [Online]. Available: \url{https://urlzs.com/2RV72}
\BIBentrySTDinterwordspacing

\bibitem{xia2018predicting}
Y.~Xia, D.~Zhang, J.~Kim, K.~Nakayama, K.~Zipser, and D.~Whitney, ``Predicting driver attention in critical situations,'' in \emph{Asian conference on computer vision}.\hskip 1em plus 0.5em minus 0.4em\relax Springer, 2018, pp. 658--674.

\bibitem{fang2019dada}
J.~Fang, D.~Yan, J.~Qiao, J.~Xue, H.~Wang, and S.~Li, ``Dada-2000: Can driving accident be predicted by driver attentionƒ analyzed by a benchmark,'' in \emph{2019 IEEE Intelligent Transportation Systems Conference (ITSC)}.\hskip 1em plus 0.5em minus 0.4em\relax IEEE, 2019, pp. 4303--4309.

\bibitem{palazzi2018predicting}
A.~Palazzi, D.~Abati, F.~Solera, R.~Cucchiara \emph{et~al.}, ``Predicting the driver's focus of attention: the dr (eye) ve project,'' \emph{IEEE transactions on pattern analysis and machine intelligence}, vol.~41, no.~7, pp. 1720--1733, 2018.

\bibitem{fang2021dada}
J.~Fang, D.~Yan, J.~Qiao, J.~Xue, and H.~Yu, ``Dada: Driver attention prediction in driving accident scenarios,'' \emph{IEEE Transactions on Intelligent Transportation Systems}, vol.~23, no.~6, pp. 4959--4971, 2021.

\bibitem{rong2022and}
Y.~Rong, N.-R. Kassautzki, W.~Fuhl, and E.~Kasneci, ``Where and what: Driver attention-based object detection,'' \emph{Proceedings of the ACM on Human-Computer Interaction}, vol.~6, no. ETRA, pp. 1--22, 2022.

\bibitem{kotseruba2022attention}
I.~Kotseruba and J.~K. Tsotsos, ``Attention for vision-based assistive and automated driving: A review of algorithms and datasets,'' \emph{IEEE Transactions on Intelligent Transportation Systems}, 2022.

\bibitem{pal2020looking}
A.~Pal, S.~Mondal, and H.~I. Christensen, ``" looking at the right stuff"-guided semantic-gaze for autonomous driving,'' in \emph{Proceedings of the IEEE/CVF Conference on Computer Vision and Pattern Recognition}, 2020, pp. 11\,883--11\,892.

\bibitem{ahlstrom2021eye}
C.~Ahlstr{\"o}m, K.~Kircher, M.~Nystr{\"o}m, and B.~Wolfe, ``Eye tracking in driver attention research—how gaze data interpretations influence what we learn,'' \emph{Frontiers in neuroergonomics}, vol.~2, p. 778043, 2021.

\bibitem{alletto2016dr}
S.~Alletto, A.~Palazzi, F.~Solera, S.~Calderara, and R.~Cucchiara, ``Dr (eye) ve: a dataset for attention-based tasks with applications to autonomous and assisted driving,'' in \emph{Proceedings of the ieee conference on computer vision and pattern recognition workshops}, 2016, pp. 54--60.

\bibitem{deng2019drivers}
T.~Deng, H.~Yan, L.~Qin, T.~Ngo, and B.~Manjunath, ``How do drivers allocate their potential attention? driving fixation prediction via convolutional neural networks,'' \emph{IEEE Transactions on Intelligent Transportation Systems}, vol.~21, no.~5, pp. 2146--2154, 2019.

\bibitem{simonyan2015deep}
K.~Simonyan and A.~Zisserman, ``Very deep convolutional networks for large-scale image recognition,'' 2015.

\bibitem{9710860}
S.~Baee, E.~Pakdamanian, I.~Kim, L.~Feng, V.~Ordonez, and L.~Barnes, ``Medirl: Predicting the visual attention of drivers via maximum entropy deep inverse reinforcement learning,'' in \emph{2021 IEEE/CVF International Conference on Computer Vision (ICCV)}, 2021, pp. 13\,158--13\,168.

\bibitem{celebi2015linear}
M.~E. Celebi and H.~A. Kingravi, ``Linear, deterministic, and order-invariant initialization methods for the k-means clustering algorithm,'' \emph{Partitional clustering algorithms}, pp. 79--98, 2015.

\bibitem{han2015deep}
S.~Han, H.~Mao, and W.~J. Dally, ``Deep compression: Compressing deep neural networks with pruning, trained quantization and huffman coding,'' \emph{arXiv preprint arXiv:1510.00149}, 2015.

\bibitem{glenn_jocher_2020_4154370}
\BIBentryALTinterwordspacing
G.~Jocher, A.~Stoken, J.~Borovec, NanoCode012, ChristopherSTAN, L.~Changyu, Laughing, tkianai, A.~Hogan, lorenzomammana, yxNONG, AlexWang1900, L.~Diaconu, Marc, wanghaoyang0106, ml5ah, Doug, F.~Ingham, Frederik, Guilhen, Hatovix, J.~Poznanski, J.~Fang, L.~Yu, changyu98, M.~Wang, N.~Gupta, O.~Akhtar, PetrDvoracek, and P.~Rai, ``{ultralytics/yolov5: v3.1 - Bug Fixes and Performance Improvements},'' Oct. 2020. [Online]. Available: \url{https://doi.org/10.5281/zenodo.4154370}
\BIBentrySTDinterwordspacing

\bibitem{bdd100k}
F.~Yu, H.~Chen, X.~Wang, W.~Xian, Y.~Chen, F.~Liu, V.~Madhavan, and T.~Darrell, ``Bdd100k: A diverse driving dataset for heterogeneous multitask learning,'' in \emph{The IEEE Conference on Computer Vision and Pattern Recognition (CVPR)}, June 2020.

\bibitem{dosovitskiy2010image}
A.~Dosovitskiy, L.~Beyer, A.~Kolesnikov, D.~Weissenborn, X.~Zhai, T.~Unterthiner, M.~Dehghani, M.~Minderer, G.~Heigold, S.~Gelly \emph{et~al.}, ``An image is worth 16x16 words: Transformers for image recognition at scale. arxiv 2020,'' \emph{arXiv preprint arXiv:2010.11929}, 2010.

\bibitem{woo2018cbam}
S.~Woo, J.~Park, J.-Y. Lee, and I.~S. Kweon, ``Cbam: Convolutional block attention module,'' in \emph{Proceedings of the European conference on computer vision (ECCV)}, 2018, pp. 3--19.

\bibitem{vaswani2017attention}
A.~Vaswani, N.~Shazeer, N.~Parmar, J.~Uszkoreit, L.~Jones, A.~N. Gomez, {\L}.~Kaiser, and I.~Polosukhin, ``Attention is all you need,'' \emph{Advances in neural information processing systems}, vol.~30, 2017.

\bibitem{NEURIPS2019_9015}
A.~Paszke, S.~Gross, F.~Massa, A.~Lerer, J.~Bradbury, G.~Chanan, T.~Killeen, Z.~Lin, N.~Gimelshein, L.~Antiga, A.~Desmaison, A.~Kopf, E.~Yang, Z.~DeVito, M.~Raison, A.~Tejani, S.~Chilamkurthy, B.~Steiner, L.~Fang, J.~Bai, and S.~Chintala, ``Pytorch: An imperative style, high-performance deep learning library,'' in \emph{Advances in Neural Information Processing Systems 32}.\hskip 1em plus 0.5em minus 0.4em\relax Curran Associates, Inc., 2019, pp. 8024--8035.

\bibitem{kingma2014adam}
D.~P. Kingma and J.~Ba, ``Adam: A method for stochastic optimization,'' \emph{arXiv preprint arXiv:1412.6980}, 2014.

\bibitem{crowston2012amazon}
K.~Crowston, ``Amazon mechanical turk: A research tool for organizations and information systems scholars,'' in \emph{Shaping the Future of ICT Research. Methods and Approaches: IFIP WG 8.2, Working Conference, Tampa, FL, USA, December 13-14, 2012. Proceedings}.\hskip 1em plus 0.5em minus 0.4em\relax Springer, 2012, pp. 210--221.

\bibitem{varol2017learning}
G.~Varol, J.~Romero, X.~Martin, N.~Mahmood, M.~J. Black, I.~Laptev, and C.~Schmid, ``Learning from synthetic humans,'' in \emph{Proceedings of the IEEE conference on computer vision and pattern recognition}, 2017, pp. 109--117.

\bibitem{peng2018visda}
X.~Peng, B.~Usman, N.~Kaushik, D.~Wang, J.~Hoffman, and K.~Saenko, ``Visda: A synthetic-to-real benchmark for visual domain adaptation,'' in \emph{Proceedings of the IEEE Conference on Computer Vision and Pattern Recognition Workshops}, 2018, pp. 2021--2026.

\bibitem{yolofast}
\BIBentryALTinterwordspacing
R.~Sovit and G.~Vikas. (2022). [Online]. Available: \url{https://learnopencv.com/performance-comparison-of-yolo-models/}
\BIBentrySTDinterwordspacing

\bibitem{yolov7}
C.~Wang, A.~Bochkovskiy, and H.~M. Liao, ``Yolov7: Trainable bag-of-freebies sets new state-of-the-art for real-time object detectors,'' in \emph{{IEEE/CVF} Conference on Computer Vision and Pattern Recognition, {CVPR} 2023, Vancouver, BC, Canada, June 17-24, 2023}.\hskip 1em plus 0.5em minus 0.4em\relax {IEEE}, 2023, pp. 7464--7475.

\end{thebibliography}
}

\end{document}